\documentclass[conference]{IEEEtran}

\usepackage{amssymb}
\usepackage{epsfig,amsmath}
\usepackage{enumerate}
\usepackage{dsfont}
\usepackage{subfigure}

\usepackage{url}

\usepackage{psfrag}
\usepackage{cite}

\makeatletter
\def\@normalsize{\@setsize\normalsize{10pt}\xpt\@xpt
\abovedisplayskip 10pt plus2pt minus5pt\belowdisplayskip \abovedisplayskip
\abovedisplayshortskip \z@ plus3pt\belowdisplayshortskip 6pt plus3pt
minus3pt\let\@listi\@listI}
\def\subsize{\@setsize\subsize{12pt}\xipt\@xipt}
\def\section{\@startsection {section}{1}{\z@}{1.0ex plus 1ex minus
 .2ex}{.2ex plus .2ex}{\large\bf}}
\def\subsection{\@startsection {subsection}{2}{\z@}{.2ex plus 1ex}
{.2ex plus .2ex}{\subsize\bf}}
\makeatother

\newtheorem{theorem}{Theorem}

\def\df{\; \stackrel \triangle = \;}

\def\adots{\mathinner{\mkern1mu\raise1pt\vbox{\kern7pt\hbox{.}}
                      \mkern2mu\raise4pt\hbox{.}
                      \mkern2mu\raise7pt\hbox{.}\mkern1mu}}

\def\R{\mbox{I{\kern-0.2em}R}}

\newcommand{\bh}{\mbox{\boldmath $h$}}

\newcommand{\by}{\mbox{\boldmath $y$}}

\newcommand{\bx}{\mbox{\boldmath $x$}}

\def\be{\begin{equation}}
\def\ee{\end{equation}}
\def\half{{\textstyle{1\over 2}}}

\def\half{{\textstyle{1\over 2}}}
\def\1overN{{\textstyle{1\over N}}}

\def\sK{{\scriptstyle {K}}}

\def\be{\setcounter{subequation}{0}
        \begin{equation}}
\def\ee{\end{equation}}

\newcounter{subequation}
\newcounter{oldsub}
\renewcommand
   {\theequation}
   {\arabic{equation}\alph{subequation}}

\def\baselabel#1
   {\setcounter{oldsub}{\value{subequation}}
    \setcounter{subequation}{0}
    \label{#1} \setcounter{subequation}{\value{oldsub}}}

\def\bses{\setcounter{subequation}{1}
          \begin{equation}}

\def\eses{\end{equation}
          \setcounter{subequation}{0}}

\def\bse{\stepcounter{subequation}
         \addtocounter{equation}{-1}
         \begin{equation}}

\def\ese{\end{equation}}

\def\refsetcounter#1#2{
   \setcounter{#1}{#2}}

\def\bsea{\begin{eqnarray}
          \refsetcounter{subequation}{1}}

\def\esea{\end{eqnarray}\refsetcounter{subequation}{0}}

\newcounter{algorithm}[section]

\def\boxit#1{\vbox{\hrule\hbox{\vrule\kern3pt
             \vbox{\kern3pt#1\kern3pt}\kern3pt\vrule}\hrule}}

\def\dspace{\multiply\normalbaselineskip 150
            \divide\normalbaselineskip 100 \normalbaselines
            \csname @@normalbaselineskip\endcsname\normalbaselineskip}

\def\sspace{\multiply\normalbaselineskip 200
            \divide\normalbaselineskip 300 \normalbaselines
            \csname @@normalbaselineskip\endcsname\normalbaselineskip}

\def\thenewbibliography#1#2{\section*{#2\markboth
 {REFERENCES}{REFERENCES}}\list
 {[\arabic{enumi}]}{\settowidth\labelwidth{[#1]}\leftmargin\labelwidth
 \advance\leftmargin\labelsep
\parsep -4pt
\usecounter{enumi}}
 \def\newblock{\hskip .11em plus .33em minus -.07em}
 \sloppy
 \sfcode`\.=1000\relax}

\def\appendix{\par
    \setcounter{section}{0} \setcounter{subsection}{0} \renewcommand
    {\theequation}
    {\Alph{section}.\arabic{equation}\alph{subequation}}
    \renewcommand{\thesection}{\Alph{section}}}

\def\R{\mbox{I{\kern-0.2em}R}}

\def\Psiit{{\mathit \Psi}}
\def\Phiit{{\mathit \Phi}}

\def\half{{\textstyle{1\over 2}}}

\def\1over4{{\textstyle{1\over 4}}}
\def\oneover8{{\textstyle{1\over 8}}}

\def\schi2{{\scriptscriptstyle {\chi^2}}}

\def\sPhi{{\scriptscriptstyle{\it \Phi}}}

\def\PSNR{\mbox{PSNR}}

\def\bh0{\mbox{\boldmath $h_0$}}

\def\bh{\mbox{\boldmath $h$}}

\def\Phiit{{\mathit \Phi}}
\def\Psiit{{\mathit \Psi}}

\def\bx{\mbox{\boldmath $x$}}

\def\bs{\mbox{\boldmath $s$}}

\def\by{\mbox{\boldmath $y$}}

\def\bx{\mbox{\boldmath $x$}}

\def\bchi{\mbox{\boldmath $\chi$}}
\newcommand{\bchitilde}{\mbox{\boldmath $\widetilde{\chi}$}}
\newcommand{\bxhat}{\mbox{\boldmath $\widehat{x}$}}
\newcommand{\bxtilde}{\mbox{\boldmath $\widetilde{x}$}}

\newcommand{\bshat}{\mbox{\boldmath $\widehat{s}$}}

\newcommand{\bzbar}{\mbox{\boldmath $\bar{z}$}}
\newcommand{\bztilde}{\mbox{\boldmath $\widetilde{z}$}}
\newcommand{\bstilde}{\mbox{\boldmath $\widetilde{s}$}}

\def\df{\; \stackrel \triangle = \;}

\def\sH{{\scriptstyle{\it H}}}
\def\sPhi{{\scriptstyle{\it \Phi}}}
\def\sM{{\scriptstyle {\rm M}}}
\def\ssM{{\scriptscriptstyle {\rm M}}}
\def\sI{{\scriptstyle {\rm I}}}
\def\ssI{{\scriptscriptstyle {\rm I}}}

\begin{document}

\title{Mask Iterative Hard Thresholding Algorithms for Sparse Image
  Reconstruction of Objects with Known Contour$^{\dagger}$}

\author{\IEEEauthorblockA{ Aleksandar Dogand\v{z}i\'c, Renliang Gu, and
    Kun Qiu} \IEEEauthorblockA{ECpE Department, Iowa State University\\
    3119 Coover Hall, Ames, IA 50011, email:
    \texttt{\{ald,renliang,kqiu\}@iastate.edu} } }

\maketitle

\begin{abstract} We develop mask iterative hard thresholding
  algorithms (mask IHT and mask DORE) for sparse image reconstruction
  of objects with known contour.  The measurements follow a noisy
  underdetermined linear model common in the compressive sampling
  literature.  Assuming that the contour of the object that we wish to
  reconstruct is \emph{known} and that the signal outside the contour
  is zero, we formulate a constrained residual squared error
  minimization problem that incorporates \emph{both} the geometric
  information (i.e.\ the knowledge of the object's contour) and the
  signal sparsity constraint. We first introduce a mask IHT method
  that aims at solving this minimization problem and guarantees
  monotonically non-increasing residual squared error for a given
  signal sparsity level. We then propose a double overrelaxation
  scheme for accelerating the convergence of the mask IHT algorithm.
  We also apply convex mask reconstruction approaches that employ a
  convex relaxation of the signal sparsity constraint.  In X-ray
  computed tomography (CT), we propose an \emph{automatic} scheme for
  extracting the convex hull of the inspected object from the measured
  sinograms; the obtained convex hull is used to capture the object
  contour information.  We compare the proposed mask reconstruction
  schemes with the existing large-scale sparse signal reconstruction
  methods via numerical simulations and demonstrate that, by
  exploiting both the geometric contour information of the underlying
  image and sparsity of its wavelet coefficients, we can reconstruct
  this image using a significantly smaller number of measurements than
  the existing methods.
\end{abstract}

\setcounter{footnote}{2}

\def\thefootnote{\fnsymbol{footnote}}

\footnotetext{This work was supported by the National Science
  Foundation under Grant CCF-0545571 and NSF Industry-University
  Cooperative Research Program, Center for Nondestructive Evaluation
  (CNDE), Iowa State University.}

\section{Introduction}
\label{Introduction}

\setcounter{footnote}{0}
\def\thefootnote{\arabic{footnote}}

Compressive sampling exploits the fact that most natural signals are
well described by only a few significant (in magnitude) coefficients
in some [e.g.\ discrete wavelet transform (DWT)] domain, where the
number of significant coefficients is much smaller than the signal
size. Therefore, for an $p \times 1$ vector $\bx$ representing the
signal and an appropriate $p \times p$ sparsifying transform matrix
$\Psiit$, we have $\bx = \Psiit \, \bs$, where $\bs = [s_1, s_2,
\ldots, s_p]^T$ is an $p \times 1$ signal transform-coefficient vector
with most elements having small magnitudes.  The idea behind
compressive sampling or compressed sensing is to \emph{sense} the
significant components of $\bs$ using a small number of linear
measurements:
\begin{equation}
\label{eq:lin_measure}
\by = \Phiit \, \bx
\end{equation}
where $\by$ is an $N \times 1$ measurement vector and $\Phiit$ is a
known $N \times p$ \emph{sampling matrix} with $N \leq p$; here, we
focus on the scenario where the measurements, signal coefficients, and
sampling and sparsifying transform matrices are real-valued. Practical
recovery algorithms, including convex relaxation, greedy pursuit, and
probabilistic methods, have been proposed to find the sparse solution
to the underdetermined system (\ref{eq:lin_measure}), see
\cite{TroppWright} for a survey.

Compressive sampling takes the advantage of the prior knowledge that
most natural signals are sparse in some transform domain.  In addition
to the signal sparsity, we use geometric constraints to enhance the
signal reconstruction performance.  In particular, we assume that the
contour of the object under inspection is \emph{known} and that the
signal outside the contour is zero.  A \emph{convex relaxation method}
was outlined in \cite{ManducaTrzaskoLi} for image reconstruction with
both sparsity and object contour information. (Note that
\cite{ManducaTrzaskoLi} does not provide sufficient information to
replicate its results and, furthermore, the method's development in
\cite[eqs.  (4)--(6)]{ManducaTrzaskoLi} clearly contains 
typos or errors.)  Here, we propose 
(i)
iterative hard thresholding and convex relaxation algorithms that
incorporate the object's contour information into the signal
reconstruction process and
(ii) an automatic scheme for extracting the convex hull of the
  inspected object (which captures the object contour
  information) from the measured X-ray computed tomography (CT)
  sinograms.

We introduce our measurement model in Section~\ref{MeasurementModel}
and the proposed iterative hard thresholding methods in
Section~\ref{MaskIHTandMaskDORE}.
Our mask convex relaxation
algorithms are described in Section~\ref{MaskConvexRelaxation}.  The
experimental results are given in Section~\ref{NumEx}.

We introduce the notation: $\| \cdot \|_p$ and ``$^T$'' denote
the $\ell_p$ norm and transpose, respectively, and the sparse
thresholding operator ${\cal T}_r(\bs)$ keeps the $r$
largest-magnitude elements of a vector $\bs$ intact and sets the rest
to zero, e.g.\ ${\cal T}_2( [0,1,-5,0,3,0]^T ) = [0,0,-5,0,3,0]^T$.
The largest singular value of a matrix $H$ is denoted by $\rho_{\sH}$
and is also known as the spectral norm of $H$. 
 Finally, $I_n$
and ${\bf 0}_{n \times 1}$ denote the identity matrix of size $n$ and
the $n \times 1$ vector of zeros, respectively.

\section{Measurement Model}
\label{MeasurementModel}

We incorporate the geometric constraints via the following signal
model: the elements of the $p \times 1$ signal vector $\bx=[x_1, x_2,
\ldots, x_p]^T$ are
\begin{equation}
\label{eq:signalmodel}
x_i = \left\{ \begin{array}{cc}   [\Psiit \, \bs]_i, & i \in \mathrm{M}
\\ 0, & i \notin \mathrm{M} \end{array} \right.
\end{equation}
for $i=1,2,\ldots,p$, where $[\Psiit \, \bs]_i$ denotes the $i$th
element of the vector $\Psiit \, \bs$, the mask $\mathrm{M}$ is the
set of $p_{\sM} \leq p$ indices corresponding to the signal elements
inside the contour of the inspected object, $\bs$ is the $p \times 1$
sparse signal transform-coefficient vector, and $\Psiit$ is the known
orthogonal sparsifying transform matrix satisfying
\begin{equation}
\label{eq:Psiorthonormal}
\Psiit \, \Psiit^T = \Psiit^T \, \Psiit = I_p.
\end{equation}
Therefore, the $p_{\sM} \times 1$ vector of signal elements
\emph{inside the mask $\mathrm{M}$} ($x_i, \, i \in \mathrm{M}$) is
$\bx_{\sM} = \Psiit_{\sM,:} \, \bs$, where the $p_{\sM} \times p$
matrix $\Psiit_{\sM,:}$ contains the $p_{\sM}$ rows of $\Psiit$ that
correspond to the signal indices within the mask $\mathrm{M}$. If the
resulting $\Psiit_{\sM,:}$ has zero columns, the elements of $\bs$
corresponding to these columns are not identifiable and are known to
be zero because they describe part of the image outside the mask
$\mathrm{M}$.  Define the set of indices $\mathrm{I}$ of nonzero
columns of $\Psiit_{\sM,:}$ containing $p_{\sI} \leq p$ elements and
the corresponding $p_{\sI} \times 1$ vector $\bs_{\sI}$ of
\emph{identifiable signal transform coefficients} under our signal
model. Then,
\begin{equation}
\label{eq:signalmodel2}
\bx_{\sM} = \Psiit_{\sM,\sI} \, \bs_{\sI}
\end{equation}
where the $p_{\sM} \times p_{\sI}$ matrix $\Psiit_{\sM,\sI}$ is the
\emph{restriction} of $\Psiit_{\sM,:}$ to the index set $\mathrm{I}$
and consists of the $p_{\sI}$ nonzero columns of
$\Psiit_{\sM,:}$. Now, the noiseless measurement equation
(\ref{eq:lin_measure}) becomes [see also (\ref{eq:signalmodel}) and
(\ref{eq:signalmodel2})]
\begin{equation}
\label{eq:lin_measure_mask}
\by = \Phiit \, \bx = \Phiit_{:,\sM} \, \Psiit_{\sM,\sI} \, \bs_{\sI}
\end{equation}
where the $N \times p_{\sM}$ matrix $\Phiit_{:,\sM}$ is the
restriction of the full sampling matrix $\Phiit$ to the mask index set
$\mathrm{M}$ and consists of the $p_{\sM}$ columns of the full
sampling matrix $\Phiit$ that correspond to the signal indices within
$\mathrm{M}$.  We now employ (\ref{eq:lin_measure_mask}) and formulate
the following constrained residual squared error minimization problem
that incorporates \emph{both} the geometric information (i.e.\ the
knowledge of the inspected object's contour) and the signal sparsity
constraint:
\begin{equation}
\label{eq:P0}
({\rm P}_0): \quad \quad 
\min_{\bs_{\sI}} \| \by - H \, \bs_{\sI} \|_2^2 \quad \mbox{subject to} \,\, \| \bs_{\sI} \|_0 \le r
\end{equation}
where $\| \bs_{\sI} \|_0$ counts the number of nonzero elements
in the vector $\bs_{\sI}$ and $H = \Phiit_{:,\sM} \,
\Psiit_{\sM,\sI}$.  We refer to $r$ as the \emph{signal sparsity
  level} and assume that it is \emph{known}.  Finding the exact
solution to (\ref{eq:P0}) involves a combinatorial search and is
therefore intractable in practice. In the following, we present greedy
iterative schemes that aim at solving (\ref{eq:P0}).

\section{Mask IHT and Mask DORE}
\label{MaskIHTandMaskDORE}

We first introduce a mask iterative hard thresholding (mask IHT) method and
then propose its double overrelaxation acceleration termed mask DORE.

Assume that the signal transform coefficient estimate
$\bs_{\sI}^{(q)}$ is available, where $q$ denotes the iteration index.
\emph{Iteration $(q+1)$} of our mask IHT scheme proceeds as follows:
\begin{equation}
  \label{eq:MaskIHT}
  \bs_{\sI}^{(q+1)} = {\cal T}_r\big(  \bs_{\sI}^{(q)} +
  \mu^{(q)} \, H^T \, ( \by - H \, \bs_{\sI}^{(q)} )
  \big)
\end{equation}
where $\mu^{(q)} > 0$ is a step size chosen to ensure monotonically
decreasing residual squared error, see also
Section~\ref{Stepsizeselection}.  Iterate until $\bs_{\sI}^{(q+1)}$
and $\bs_{\sI}^{(q)}$ do not differ significantly.  Upon convergence
of this iteration yielding $\bs_{\sI}^{(+\infty)}$, construct an
estimate of the signal vector $\bx_{\sM}$ inside the mask $\mathrm{M}$
using $\Psiit_{\sM,\sI} \, \bs_{\sI}^{(+\infty)}$.  In
\cite{DogandzicGuQiuQNDE11}, we consider (\ref{eq:MaskIHT}) with
constant $\mu^{(q)}$ (not a function of $q$) set to $\mu^{(q)} =
1/\rho_{\sPhi}^2$.  For the full mask $\mathrm{M}=\{1,2,\ldots,p\}$
and constant $\mu^{(q)}$, (\ref{eq:MaskIHT}) reduces to the standard
iterative hard thresholding (IHT) algorithm in
\cite{BlumensathDavies}.

We now propose our mask DORE iteration that applies \emph{two
  consecutive overrelaxation steps} after one mask IHT step to
accelerate the convergence of the mask IHT algorithm. These two
overrelaxations use the identifiable signal coefficient estimates
$\bs_{\sI}^{(q)}$ and $\bs_{\sI}^{(q-1)}$ from the two most recently
completed mask DORE iterations. \emph{Iteration $(q+1)$} of our mask
DORE scheme proceeds as follows:

\noindent
{\bf 1.\ Mask IHT step.} 
\begin{equation}
\label{eq:shat}
  \bshat_{\sI} = \bshat_{\sI}(\bs_{\sI}^{(q)},  \mu^{(q)})  = {\cal T}_r\big(  \bs_{\sI}^{(q)} + \mu^{(q)} \, H^T \, ( \by - H \, \bs_{\sI}^{(q)} )
  \big)
\end{equation}
where $\mu^{(q)} > 0$ is a step size chosen to ensure monotonically
decreasing residual squared error, see also
Section~\ref{Stepsizeselection}.

\noindent
{\bf 2.\ First overrelaxation.} 
Minimize the residual squared error $\| \by - H \, \bs_{\sI} \|_2^2$
with respect to $\bs_{\sI}$ lying on the straight line connecting $\bshat_{\sI}$ and
$\bs_{\sI}^{(q)}$:
\begin{subequations}
\begin{equation}
  \bzbar_{\sI} = \bshat_{\sI} + \alpha_1 \, (   \bshat_{\sI} - \bs_{\sI}^{(q)})
\end{equation}
which has a \emph{closed-form} solution:
\begin{equation}
\alpha_1 = \frac{( H \, \bshat_{\sI} -
 H \, \bs_{\sI}^{(q)})^T \, (\by - H \, \bshat_{\sI})}{\| H \,
  \bshat_{\sI} - H \, \bs_{\sI}^{(q)} \|_2^2}.
\end{equation}
\end{subequations}

\noindent {\bf 3.\ Second overrelaxation.}
Minimize the residual squared error $\| \by - H \, \bs_{\sI} \|_2^2$
with respect to $\bs_{\sI}$ lying on
the straight line connecting $\bzbar_{\sI}$ and $\bs_{\sI}^{(q-1)}$:
\begin{subequations}
\begin{equation}
\label{eq:ztilde}
\bztilde_{\sI} = \bzbar_{\sI} + \alpha_2 \, ( \bzbar_{\sI} - \bs_{\sI}^{(q-1)} )
\end{equation}
which has a closed-form solution:
\begin{equation}
\alpha_2 = \frac{( H \, \bzbar_{\sI} -
H \, \bs_{\sI}^{(q-1)})^T \, (\by -
 H \, \bzbar_{\sI})}{\| H \, \bzbar_{\sI} - H \, \bs_{\sI}^{(q-1)} \|_2^2}.
\end{equation}
\end{subequations}

\noindent
{\bf 4.\ Thresholding.} Threshold $\bztilde_{\sI}$ to
the sparsity level $r$: $\bstilde_{\sI} = {\cal T}_r (\bztilde_{\sI})$.

\noindent
{\bf 5.\ Decision.}
 If $\| \by - H \, \bstilde_{\sI} \|_2^2  < \| \by - H \, \bshat_{\sI} \|_2^2$,
assign $\bs_{\sI}^{(q+1)}=\bstilde_{\sI}$; otherwise, assign
$\bs_{\sI}^{(q+1)}=\bshat_{\sI}$ and complete {\em Iteration
  $q+1$}\/.

Iterate until $\bs_{\sI}^{(q+1)}$ and $\bs_{\sI}^{(q)}$ do not differ
significantly.  As before, upon convergence of this iteration yielding
$\bs_{\sI}^{(+\infty)}$, construct an estimate of the signal vector
$\bx_{\sM}$ inside the mask $\mathrm{M}$ using $\Psiit_{\sM,\sI} \,
\bs_{\sI}^{(+\infty)}$.

\subsection{Step size selection}
\label{Stepsizeselection}

In \emph{Iteration 1} of our mask DORE and mask IHT schemes, we seek the
largest step size $\mu^{(0)}$ that satisfies 
\begin{equation}
  \label{eq:stepsizeinitialcondition}
\| \by - H \, \widehat{\bs}_{\sI} \|_2^2
\leq \| \by - H \, \bs_{\sI}^{(0)} \|_2^2
\end{equation}
where $\widehat{\bs}_{\sI} = \bshat_{\sI}(\bs_{\sI}^{(0)}, \mu^{(0)})$
is computed using (\ref{eq:shat}) with $q=0$.  We achieve this goal
approximately as follows: Start with an initial guess for $\mu^{(0)} >
0$, compute the corresponding $\widehat{\bs}_{\sI} =
\bshat_{\sI}(\bs_{\sI}^{(0)}, \mu^{(0)})$, and
\begin{itemize}
\item if (\ref{eq:stepsizeinitialcondition}) holds for the initial
  step size guess, double (repeatedly, if needed) $\mu^{(0)}$ until
  the condition (\ref{eq:stepsizeinitialcondition}) for the
  corresponding $\widehat{\bs}_{\sI} = \bshat_{\sI}(\bs_{\sI}^{(0)},
  \mu^{(0)})$ fails;
\item shrink (repeatedly, if needed) $\mu^{(0)}$ by multiplying it
  with $0.9$ until (\ref{eq:stepsizeinitialcondition}) for the
  corresponding $\widehat{\bs}_{\sI} = \bshat_{\sI}(\bs_{\sI}^{(0)},
  \mu^{(0)})$ holds;
\item complete \emph{Iteration 1} by moving on to Steps 2--5 in mask
  DORE or setting $\bs_{\sI}^{(q+1)} = \widehat{\bs}_{\sI}$ in mask
  IHT.
\end{itemize}
In each subsequent \emph{Iteration $q+1$} ($q > 0$), start with
$\mu^{(q)} = \mu^{(q-1)}$, compute the corresponding
$\widehat{\bs}_{\sI} = \bshat_{\sI}(\bs_{\sI}^{(q)},  \mu^{(q)})$ in (\ref{eq:shat}), and
\begin{itemize}
\item if 
\begin{equation}
  \label{eq:stepsizecondition}
\| \by - H \, \widehat{\bs}_{\sI} \|_2^2
\leq  \| \by - H \, \bs_{\sI}^{(q)} \|_2^2
\end{equation}
 does not hold for the initial step size
$\mu^{(q)} = \mu^{(q-1)}$, shrink $\mu^{(q)}$ by multiplying it
(repeatedly, if needed) with $0.9$ until (\ref{eq:stepsizecondition})
for the corresponding $\widehat{\bs}_{\sI} = \bshat_{\sI}(\bs_{\sI}^{(q)},  \mu^{(q)})$ holds; 
\item complete \emph{Iteration $q+1$} by moving on to Steps 2--5 in
  mask DORE or setting $\bs_{\sI}^{(q+1)} = \widehat{\bs}_{\sI}$ in
  mask IHT.
\end{itemize}
Therefore, our step size $\mu^{(q)}$ is a decreasing piecewise
constant function of the iteration index $q$.  The step size
$\mu^{(+\infty)}$ obtained upon convergence (i.e.\ as $q \nearrow
+\infty$) is larger than or equal to $0.9/\rho_{\sH}^2$, which follows
easily from Theorem~\ref{theorem} below.

\begin{theorem}
  \label{theorem}
  Assuming that
\begin{equation}
  \label{eq:sufficientconditionformu}
0 < \mu^{(q)} \leq 1/\rho_{\sH}^2
\end{equation}
and that the signal coefficient estimate in the $q$-th iteration
$\bs_{\sI}^{(q)}$ belongs to the parameter space
\begin{equation}
  \label{eq:signalparameterspace}
  {\cal S}_r = \{ \bs_{\sI} \in \mathds{R}^{p_{\ssI}}: \,
  \| \bs \|_0 \leq r \, \}
\end{equation}
then (\ref{eq:stepsizecondition}) holds, where
$\widehat{\bs}_{\sI} = \bshat_{\sI}(\bs_{\sI}^{(q)}, \mu^{(q)})$ in
(\ref{eq:stepsizecondition}) is computed using
(\ref{eq:shat}). 
Consequently, under the above conditions, the mask IHT and mask
DORE iterations yield convergent monotonically nonincreasing squared residuals
$\| \by - H \, \bs_{\sI}^{(q)} \|_2^2$ as the iteration index $q$ goes
to infinity.
\end{theorem}
\begin{IEEEproof}
  See the Appendix.
\end{IEEEproof}

\section{Mask Convex Relaxation Methods}
\label{MaskConvexRelaxation}

Consider a Lagrange-multiplier formulation of (\ref{eq:P0}) with the
$\ell_0$ norm replaced by the $\ell_1$ norm:
\begin{equation}
\label{eq:P1BPDN}
({\rm P}_1): \quad \quad 
\min_{\bs_{\sI}} (\half \, \| \by - H \, \bs_{\sI}
\|_2^2 + \tau \, \| \bs_{\sI} \|_1)
\end{equation}
where $\tau$ is the regularization parameter that controls the signal
sparsity; note that the convex problem (\ref{eq:P1BPDN}) can be solved
in polynomial time. Here, we solve (\ref{eq:P1BPDN}) using the
fixed-point continuation active set (FPC$_{\footnotesize \mbox{AS}}$)
and gradient-projection for sparse reconstruction with debiasing
methods in \cite{WenYinGoldfarbZhang} and
\cite{FigueiredoNowakWright}, respectively.  We refer to these methods
as \emph{mask FPC$_{\footnotesize \mbox{AS}}$} and \emph{mask GPSR},
respectively.

\section{Automatic Mask Generation from X-ray CT Sinograms Using a
  Convex Hull of the Object}
\label{AutomaticConvexHullExtraction}

In X-ray computed tomography (CT), accurate object contour information
can be extracted \emph{automatically} from the measured sinograms. In
particular, we construct a convex hull of the inspected object by
taking intersection of the supports of the projections (over all
projection angles) in the spatial image domain. 

\begin{figure}
\centering
\includegraphics[width=2.5in]{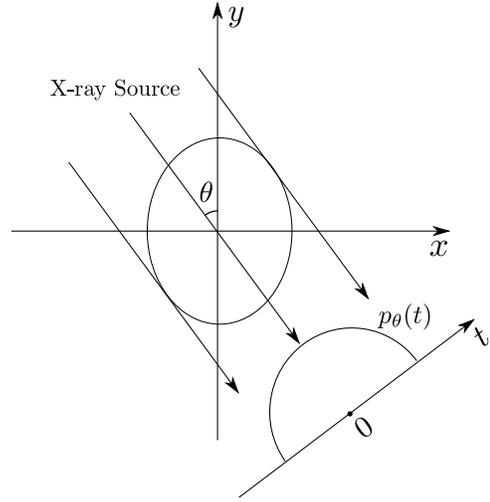}
\label{fig:sinogramconvexhull}

\caption{Geometry of the parallel-beam X-ray CT system.}
\end{figure}

To illustrate the convex hull extraction procedure, consider a
parallel-beam X-ray CT system. Denote the measured sinogram by
$p_{\theta}(t)$, where $\theta$ is the projection angle and $t$ is the
distance from the rotation center $O$ to the measurement point. To
obtain sufficient data for reconstruction, the range of $t$ must be
sufficiently large so that both ends of every projection
$p_{\theta}(t)$ are zero. Define the range of the sinogram at
angle $\theta$ by
$[a_{\theta}, b_{\theta}]=\inf\left\{[a,b]\in \mathds{R}: p_{\theta}(t) = 0 
\text{ for all } t \notin [a,b]\right\}$
and the corresponding range in the spatial image domain:
\[
A_{\theta}=\left\{(x,y)\in \mathds{R}^2: x\cos \theta+y\sin \theta \in 
[a_{\theta}, b_{\theta}]\right\}
\]
We construct the convex hull of the inspected object by
taking the intersection $\bigcap_{\theta=0}^{\pi} A_{\theta}$.
In practice, only a finite number $K$ of projections is available at
angles $\theta_1,\theta_2,\ldots,\theta_{\sK} \in [0,\pi)$, and the
corresponding convex hull of the object can be computed as
$\bigcap_{k=1}^K A_{\theta_k}$. Clearly, the angles
$\theta_1,\theta_2,\ldots,\theta_{\sK}$ determine the tightness of the
obtained convex hull.

When imaging objects whose mass density is relatively high compared
with that of the air, it is easy to determine the supports of the
projections from the measured sinograms and extract the corresponding
convex hull. For low-density objects such as pieces of foam, we need
to choose carefully a threshold for determining these supports.

\section{Numerical Examples}
\label{NumEx}

In the following examples, we use the standard \emph{filtered
  backprojection} (FBP) method \cite[Sec.\ 3.3]{KakSlaney}, which
ignores both the signal sparsity and geometric object contour
information, to initialize all iterative signal reconstruction
methods. The mask DORE and DORE methods employ the following
convergence criteria:
\begin{equation}
\label{eq:convcrit}
\| \bs_{\sI}^{(p+1)} - {\bs_{\sI}}^{(p)} \|_2^2 \big/ p_{\sI} < \epsilon,
\quad
\| \bs^{(p+1)} - \bs^{(p)} \|_2^2 \big/ p < \epsilon
\end{equation}
respectively, where $\epsilon > 0$ denotes the convergence threshold.

\begin{figure*}[!t]
\centering
\subfigure[]{
\includegraphics[width=2.2in]{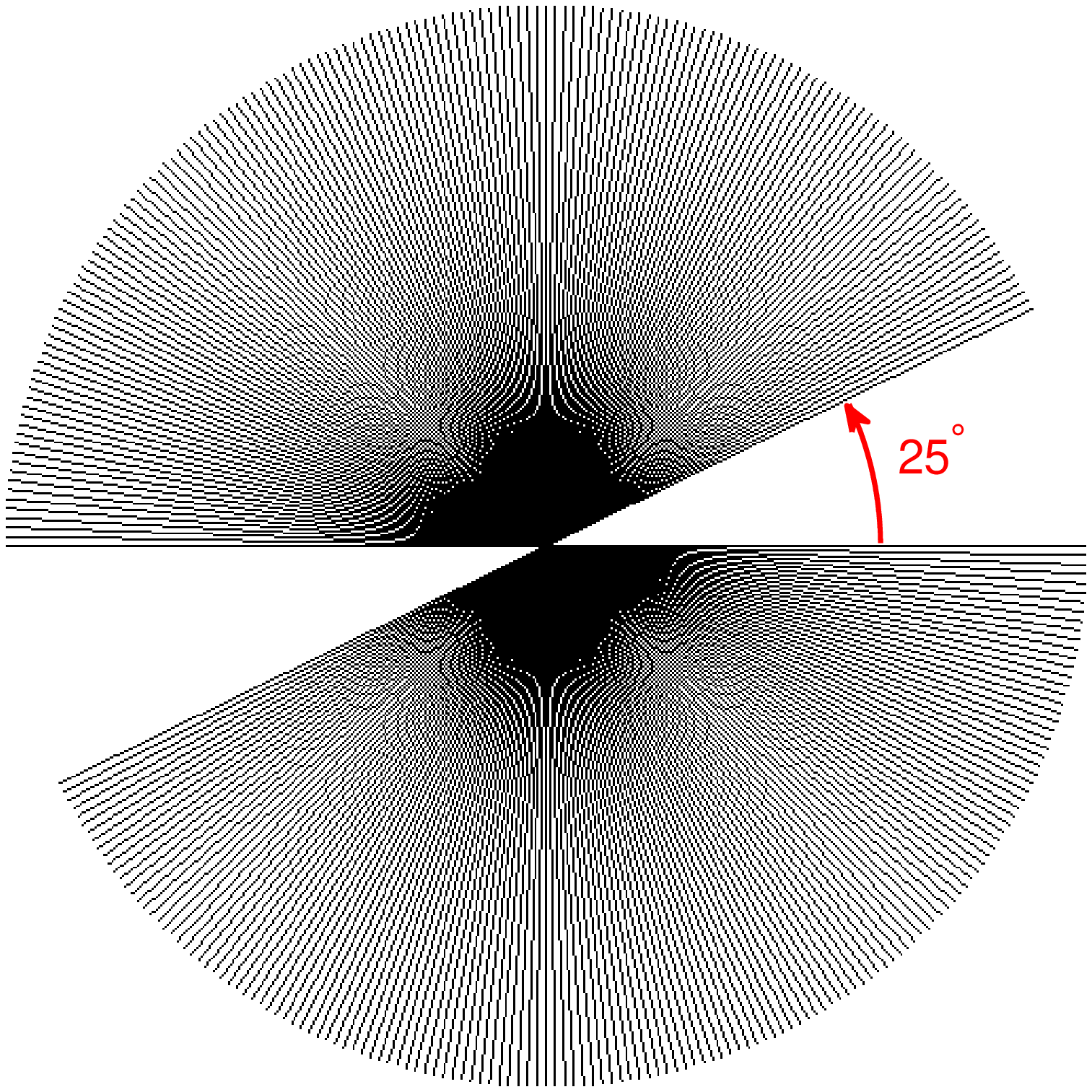}
\label{fig:phantom_sample}
}
\hfil
\subfigure[]{
\includegraphics[width=2.2in]{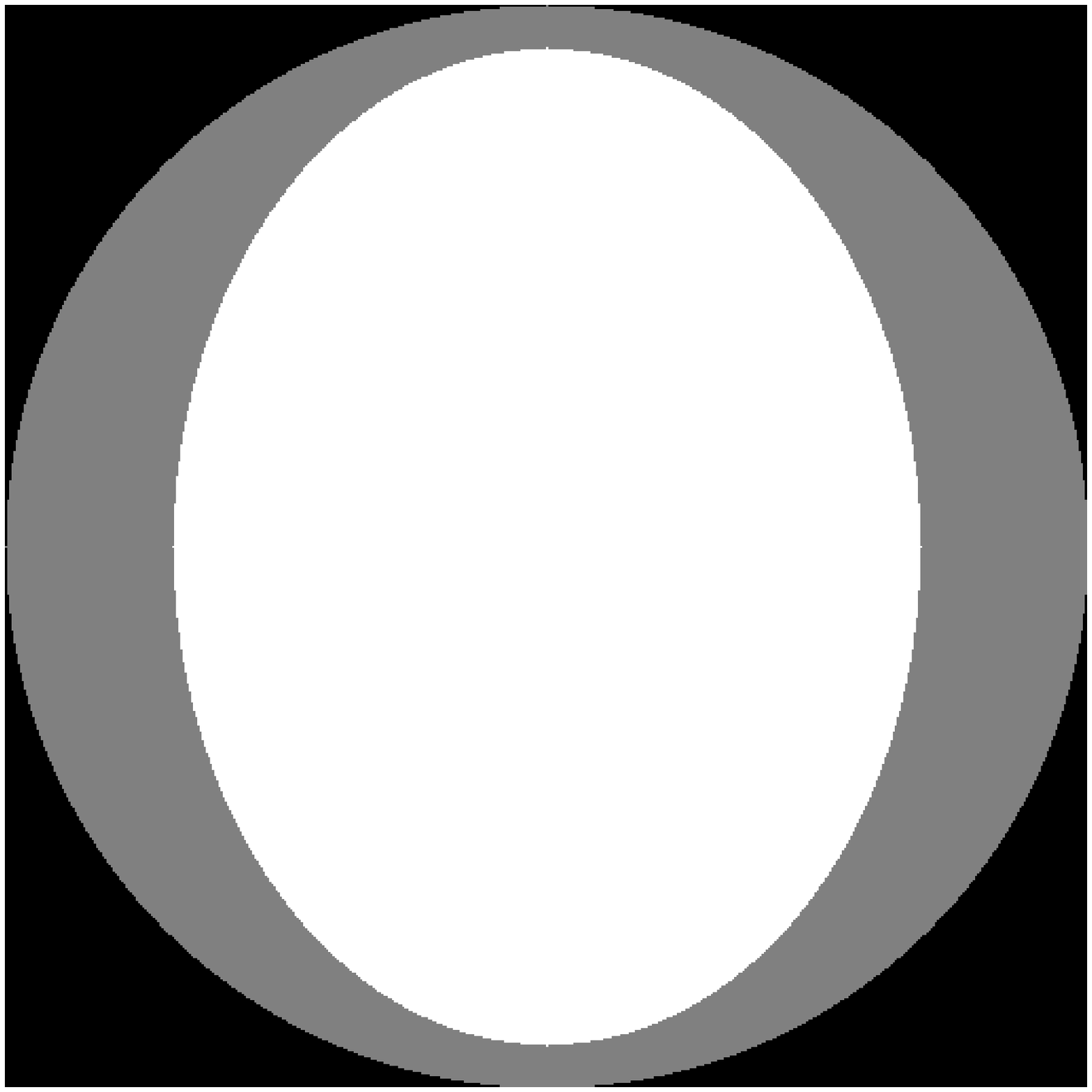}
\label{fig:phantom_mask}
}
\hfil
\subfigure[]{
\includegraphics[width=2.2in]{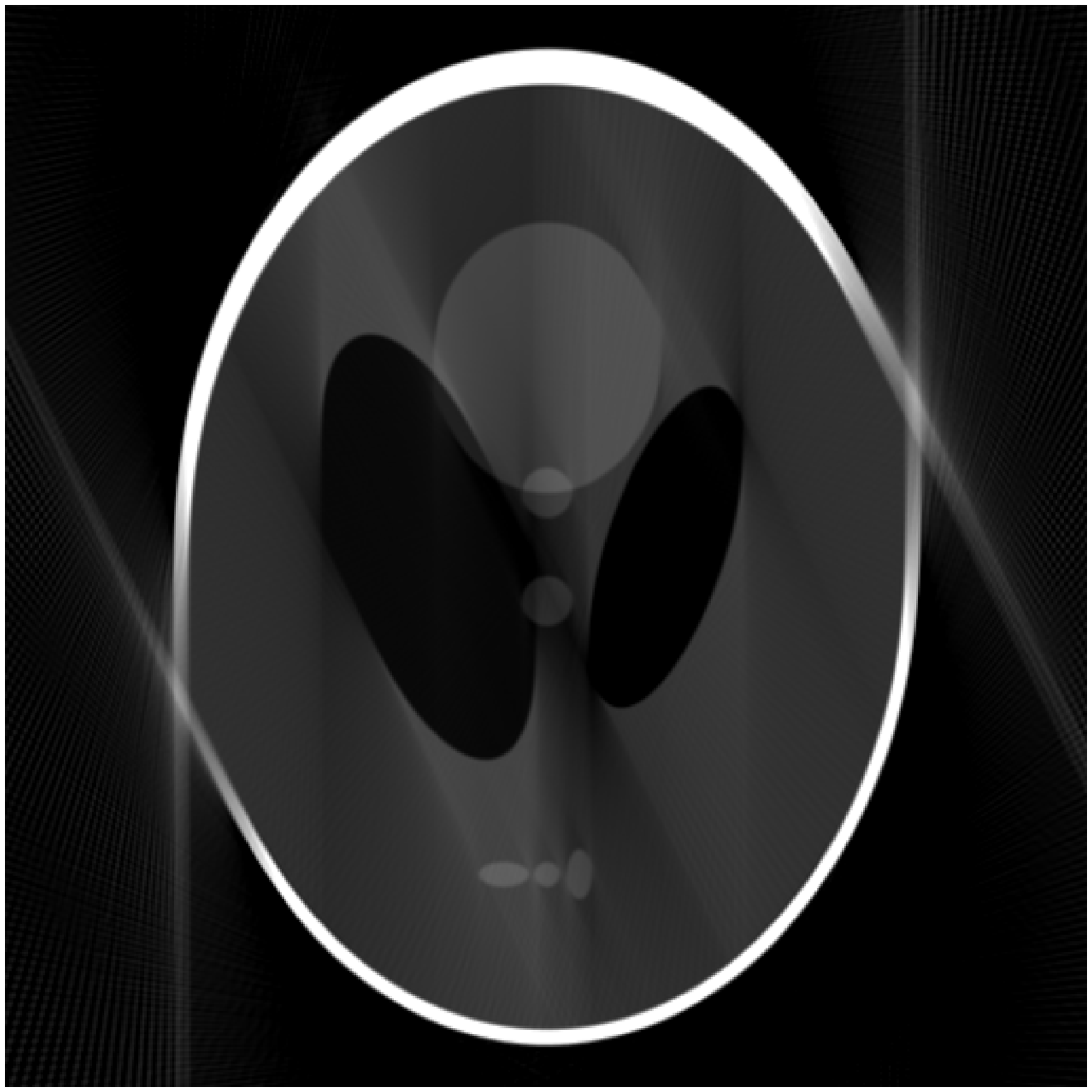}
\label{fig:phantom_FBP}
}
\hfil
\subfigure[]{
\includegraphics[width=2.2in]{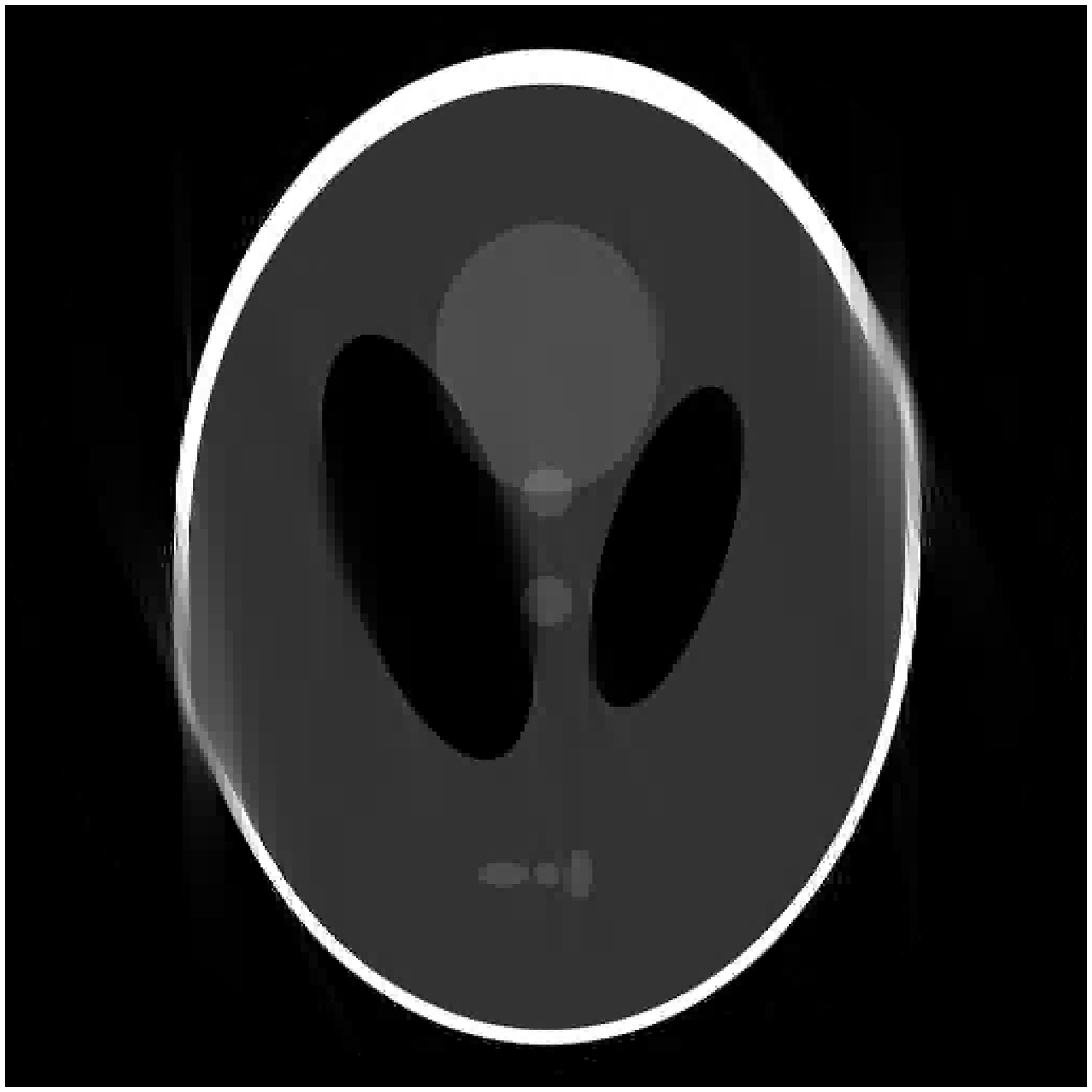}
\label{fig:phantom_DORE}
}
\hfil
\subfigure[]{
\includegraphics[width=2.2in]{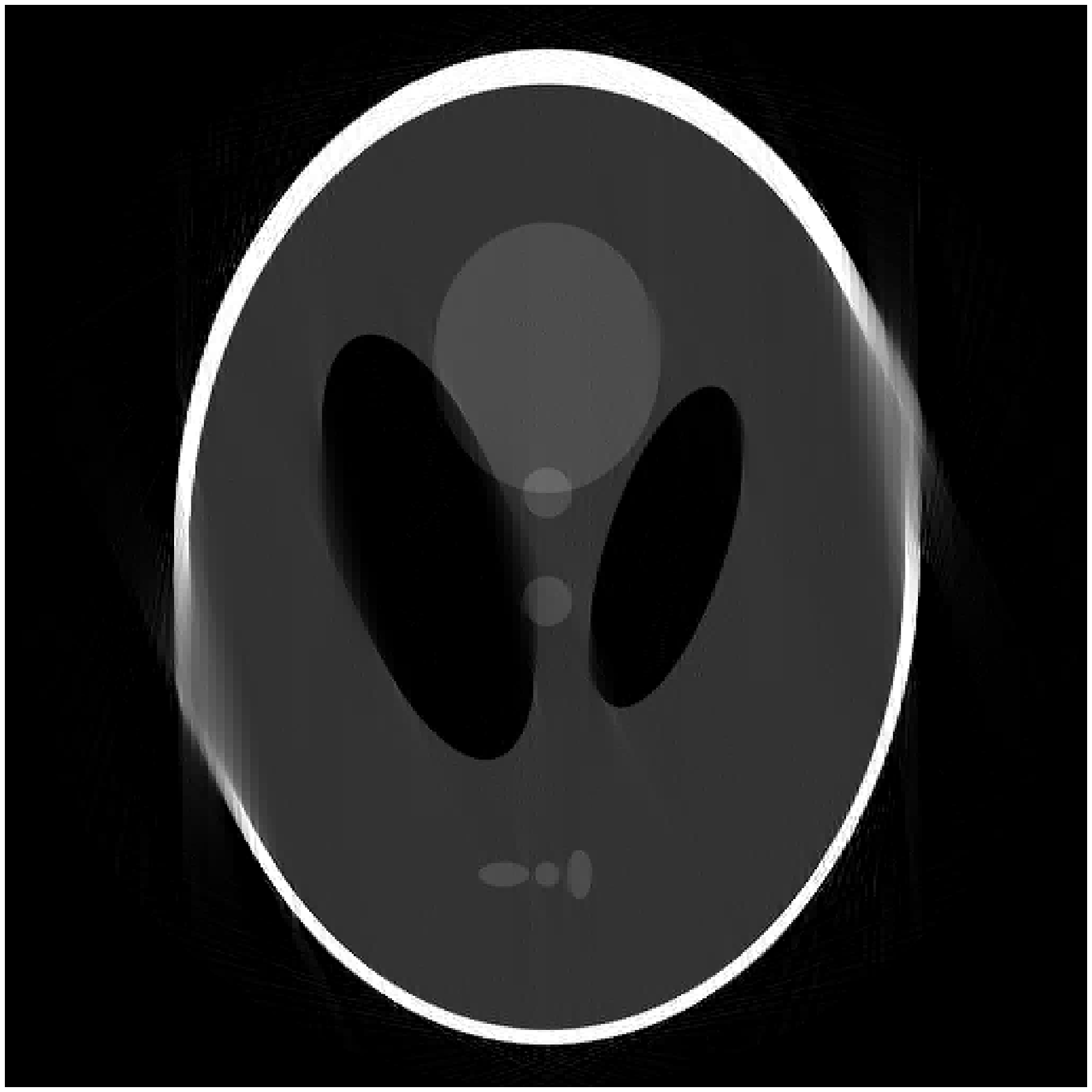}
\label{fig:phantom_GPSR}
}
\hfil
\subfigure[]{
\includegraphics[width=2.2in]{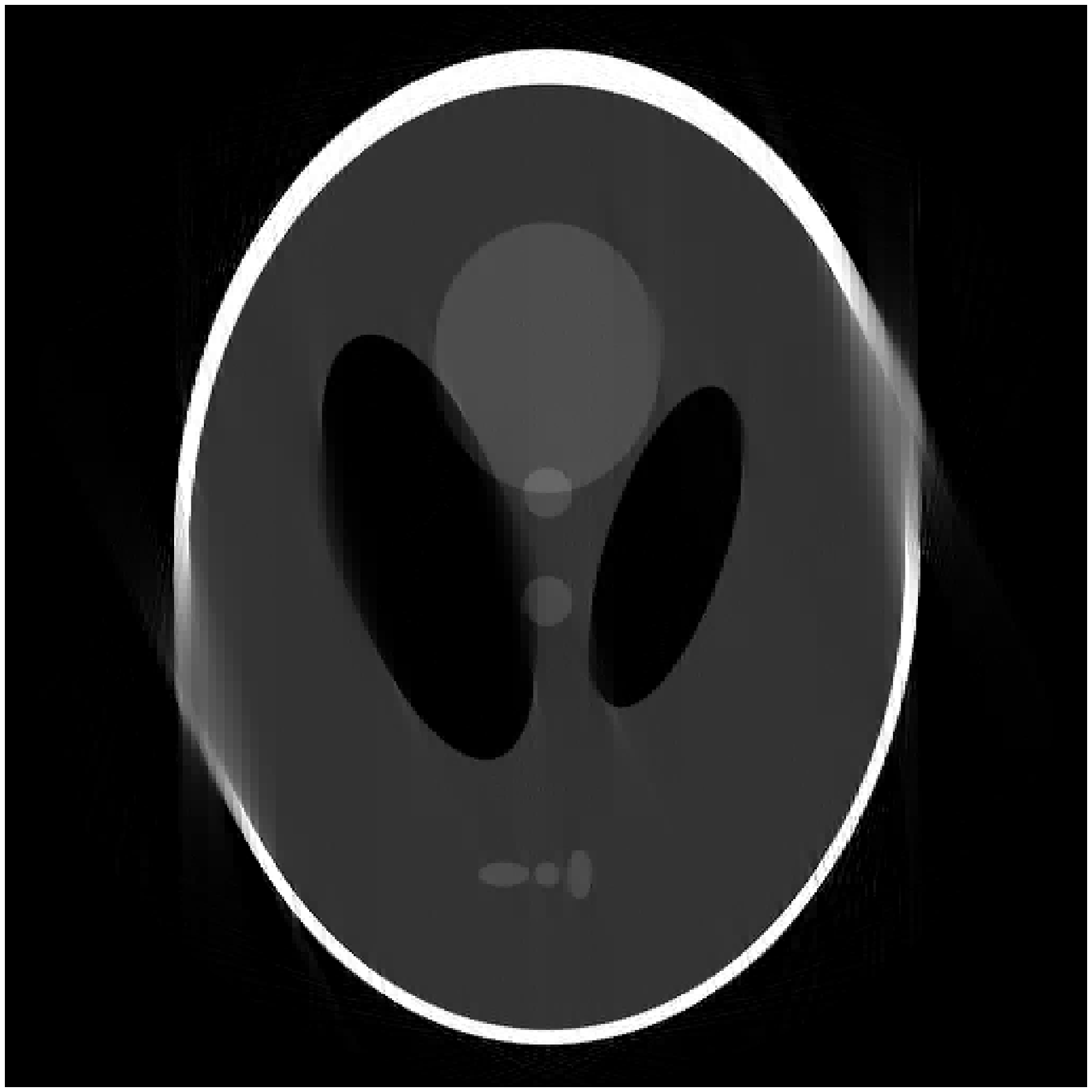}
\label{fig:phantom_FPC}
}
\hfil
\subfigure[]{
\includegraphics[width=2.2in]{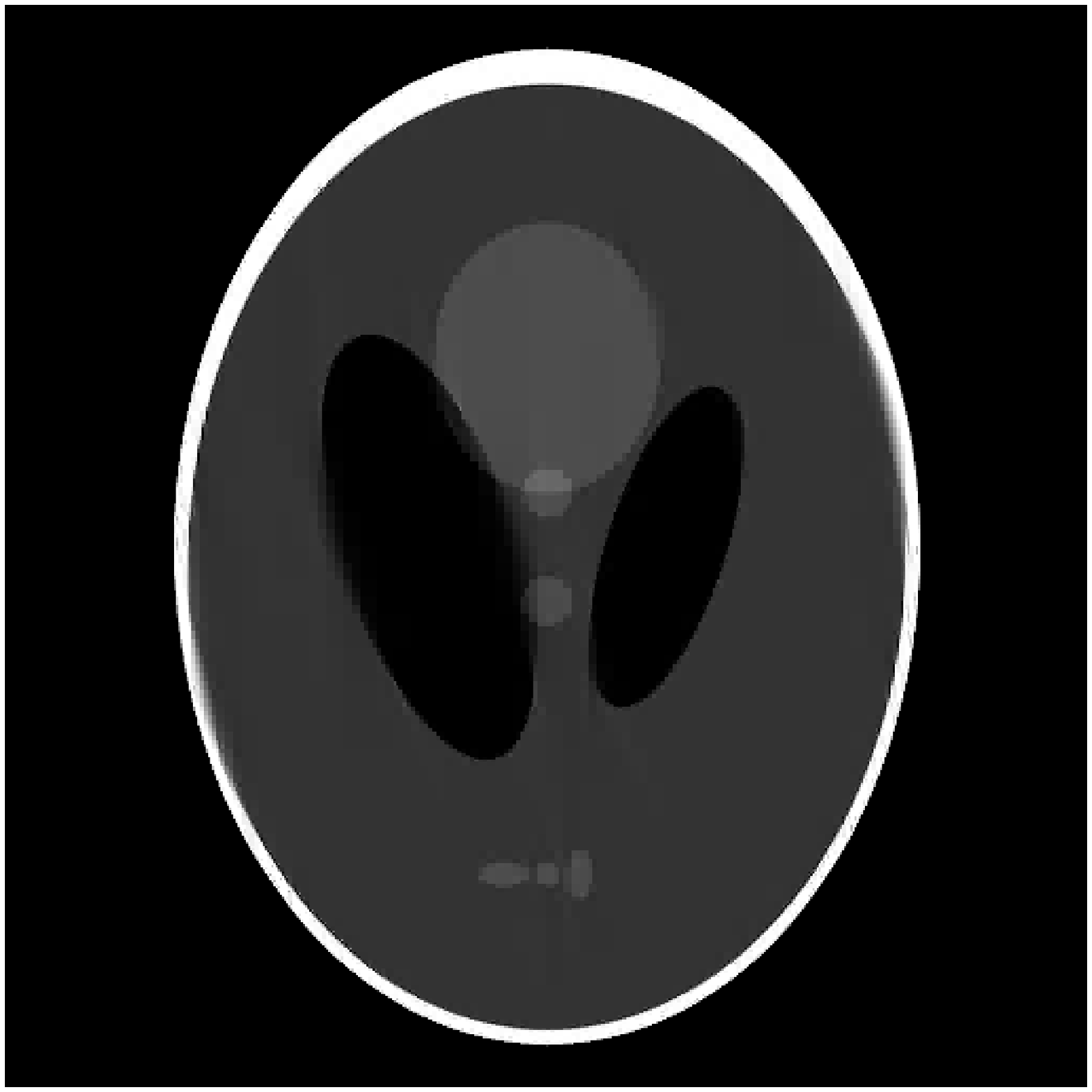}
\label{fig:phantom_maskDORE}
}
\hfil
\subfigure[]{
\includegraphics[width=2.2in]{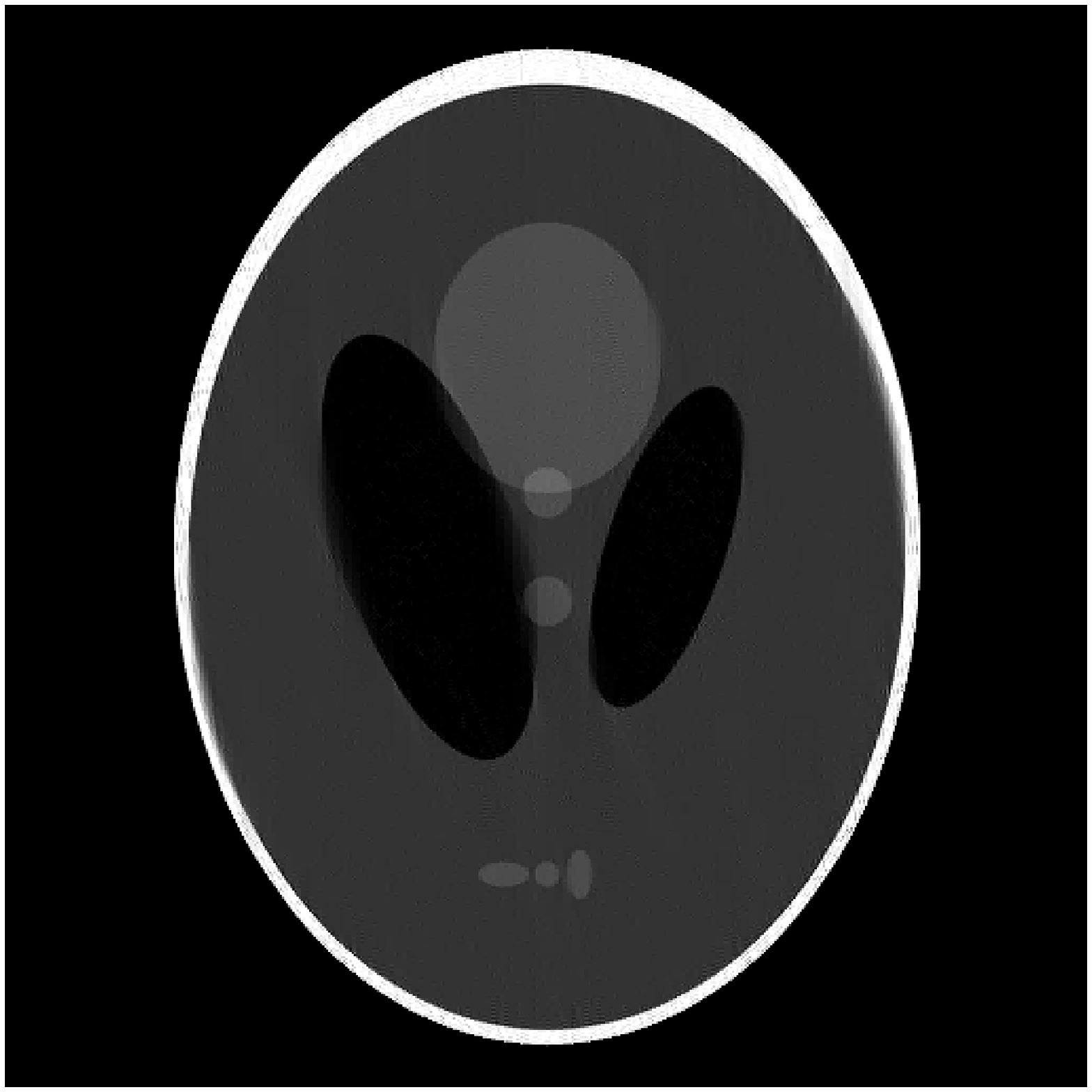}
\label{fig:phantom_maskGPSR}
}
\hfil
\subfigure[]{
\includegraphics[width=2.2in]{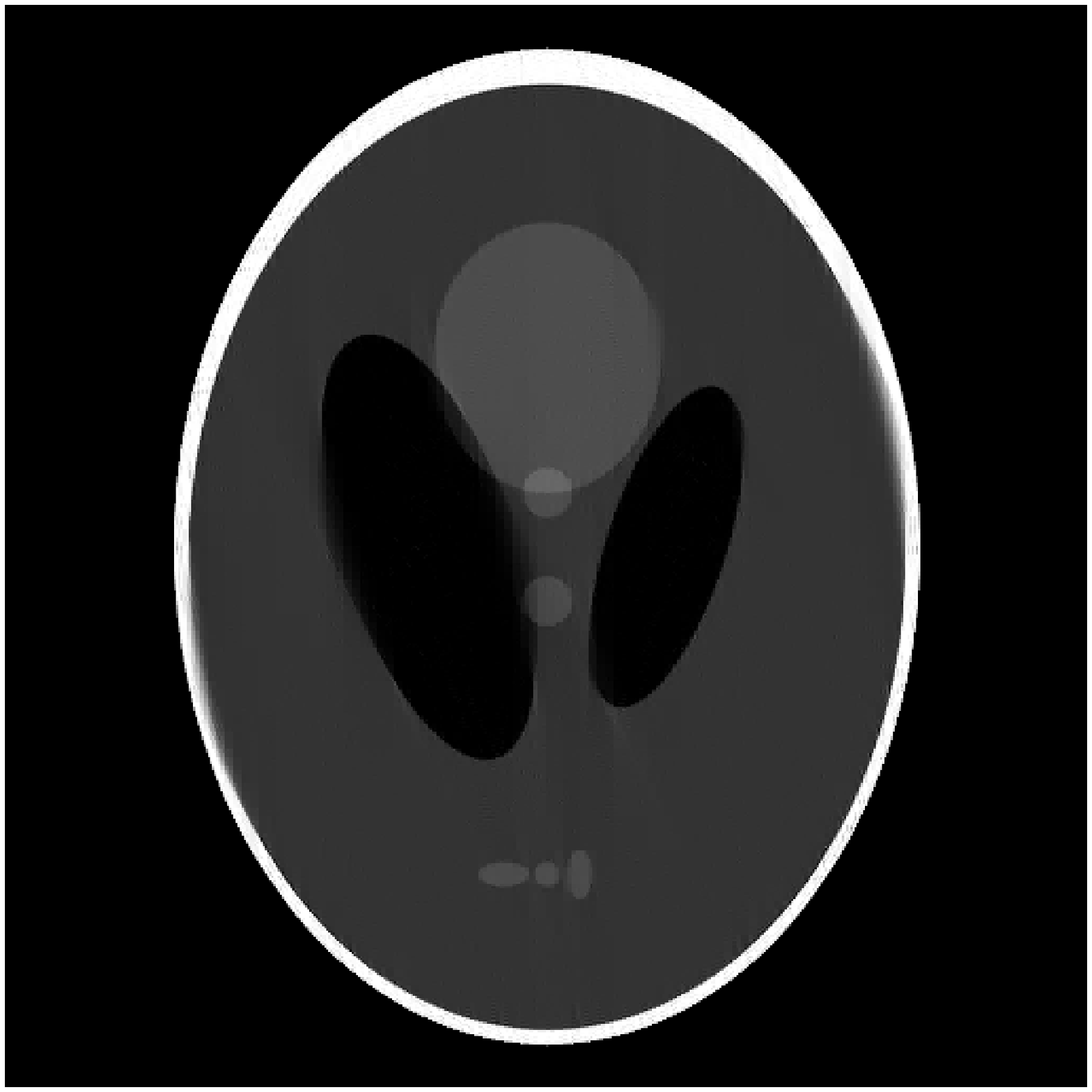}
\label{fig:phantom_maskFPCAS}
}

\caption{(a) $155$ limited-angle projections in the 2-D frequency
  plane, (b) the full  and outer-shell masks of the 
  Shepp-Logan phantom, (c) FBP ($\PSNR=19.9$~dB), (d) DORE
  ($\PSNR=22.7$~dB), (e) GPSR ($\PSNR=22.9$~dB), (f)
  FPC$_{\footnotesize \mbox{AS}}$ ($\PSNR=22.5$~dB), (g) mask DORE
  ($\PSNR=25.8$~dB), (h) mask GPSR ($\PSNR=25.3$~dB), and (i) mask
  FPC$_{\footnotesize \mbox{AS}}$ ($\PSNR=26.4$~dB) reconstructions.}
\label{fig:numex1}
\end{figure*}

\textbf{\textit{Shepp-Logan phantom reconstruction.}}  We simulated
limited-angle parallel-beam projections of an
\emph{analog}
Shepp-Logan phantom with $1^\circ$ spacing between projections and
missing angle span of $25^\circ$. Each projection is computed from its
analytical sinogram using \cite[function
\texttt{ellipse\_sino.m}]{Fessler} and \cite{KakSlaney} and then
sampled by a receiver array containing $511$ elements.
We then compute FFT of each projection, yielding $N=512$
frequency-domain measurements; the corresponding frequency-domain
sampling pattern is shown in Fig.~\ref{fig:phantom_sample}.

Fig.~\ref{fig:phantom_mask} depicts both the full and outer-shell
masks of the phantom that we use to implement the DORE, GPSR,
FPC$_{\footnotesize \mbox{AS}}$ and mask DORE, GPSR, and
FPC$_{\footnotesize \mbox{AS}}$ methods, respectively. Because of the
nature of X-ray CT measurements, our full mask has circular shape
containing $p=205859$ signal elements. The elliptical outer-shell mask
containing $p_{\sM} = 130815 \approx 0.6355 \,p$ pixels has been
constructed from the phantom's sinogram using $\bigcap_{k=1}^{180} A_{
  \pi \, (k-1) / 180 }$, see
Section~\ref{AutomaticConvexHullExtraction}; this choice of the mask
implies that we have prior information about the shape of the outer
shell of the Shepp-Logan phantom beyond the information available from
the limited-angle projections that we use for reconstruction, see
Fig.~\ref{fig:phantom_sample}. \looseness=-1

Our performance metric is the peak signal-to-noise ratio (PSNR) of a
reconstructed image $\widehat{\bx} = [\widehat{x}_1, \widehat{x}_2,
\ldots, \widehat{x}_p]^T$ inside the mask M:
\[
\PSNR~\mbox{(dB)} =  10 \, \log_{10} \Big\{ \frac{ [(\max_{i \in \sM} x_i) - (\min_{i
    \in \sM} x_i)]^2}{ \sum_{i \in \sM} (\widehat{x}_i - x_i)^2  / p_{\sM}}  \Big\}
\]
where $\bx$ is the true image.

 We select the inverse Haar (Daubechies-2) DWT matrix to be the
  orthogonal sparsifying transform matrix $\Psiit$; the true signal
  vector $\bs$ consists of the Haar wavelet transform coefficients of
  the phantom and is sparse:
\[
\| \bs \|_0 = 7866 \approx 0.0382 \, p.
\]
For the above choices of the mask and sparsifying transform, the
number of identifiable signal transform coefficients is $p_{\sI} =
132450 \approx 0.6434 \, p$. Note that $\| \bs \|_0 = \| \bs_{\sI}
\|_0 \ll p_{\sI}$, implying that the identifiable signal coefficients
are sparse as well.

We compare the reconstruction performances of
\begin{itemize}
\item mask DORE ($r=7000$) and DORE ($r=8000$) with
  $\epsilon=10^{-14}$ [see (\ref{eq:convcrit})], where $r$ are tuned for good PSNR performance;
\item the mask FPC$_{\footnotesize \mbox{AS}}$, mask GPSR,
  FPC$_{\footnotesize \mbox{AS}}$, and GPSR schemes,
 all using the
  regularization parameter $\tau=10^{-5} \, \| H^T \, \by
  \|_{\infty}$ tuned for good PSNR performance;
\item the standard FBP method.
\end{itemize}
(Here, we
    employ the convergence threshold
  $\texttt{tolP} = 10^{-5}$ for the mask GPSR and GPSR schemes, see \cite{FigueiredoNowakWright}.)

Figs.~\ref{fig:phantom_FBP}--\ref{fig:phantom_maskFPCAS} show the
reconstructions of various methods. To facilitate comparison, we
employ the common gray scale to represent the pixel values within the
images in Figs.~\ref{fig:phantom_FBP}--\ref{fig:phantom_maskFPCAS}.
Clearly, taking the object's contour into account improves the 
signal reconstruction performance.


\textbf{\textit{Industrial object reconstruction.}}  We apply our
proposed methods to reconstruct an industrial object from real
fan-beam X-ray CT projections.  First, we performed the standard
fan-to-parallel beam conversion (see \cite[Sec.\ 3.4]{KakSlaney}) and
generated parallel-beam projections with $1^\circ$ spacing and
measurement array size of $1023$ elements, yielding $N=1024$
frequency-domain measurements per projection.  Our full mask has circular shape
containing $p=823519$ signal elements. The outer-shell mask containing
$p_{\sM} = 529079 \approx 0.6425 \,p$ pixels has been constructed from the
phantom's parallel-beam sinogram using $\bigcap_{k=1}^{180} A_{ \pi \,
  (k-1) / 180 }$, see Section~\ref{AutomaticConvexHullExtraction}.

\begin{figure*}[!t]
\centering
\subfigure[]{
\includegraphics[width=2.2in]{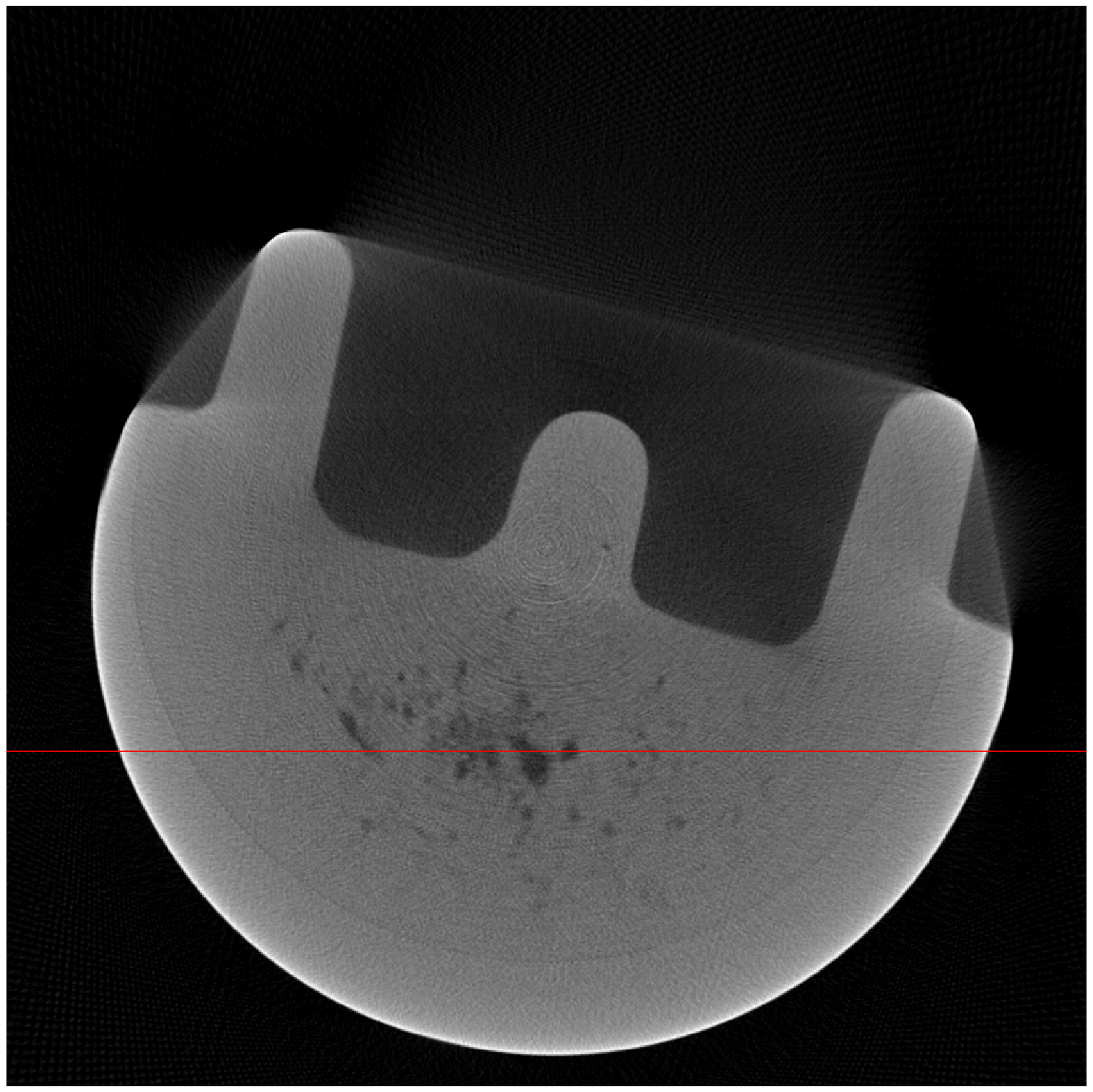}
\label{fig:img_FBPfull}
}
\hfil
\subfigure[]{
\includegraphics[width=2.2in]{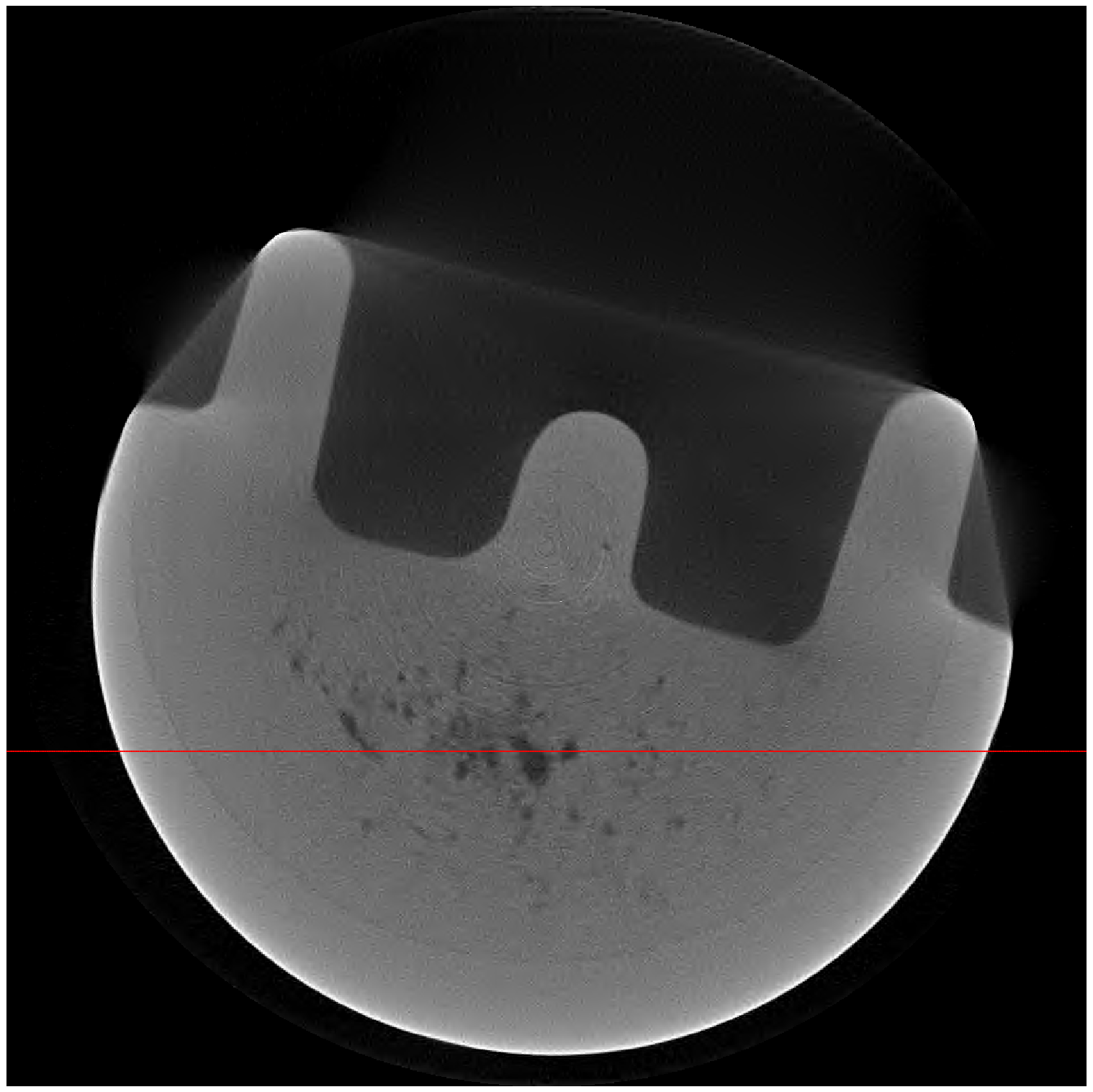}
\label{fig:img_DOREfull}
}
\hfil
\subfigure[]{
\includegraphics[width=2.2in]{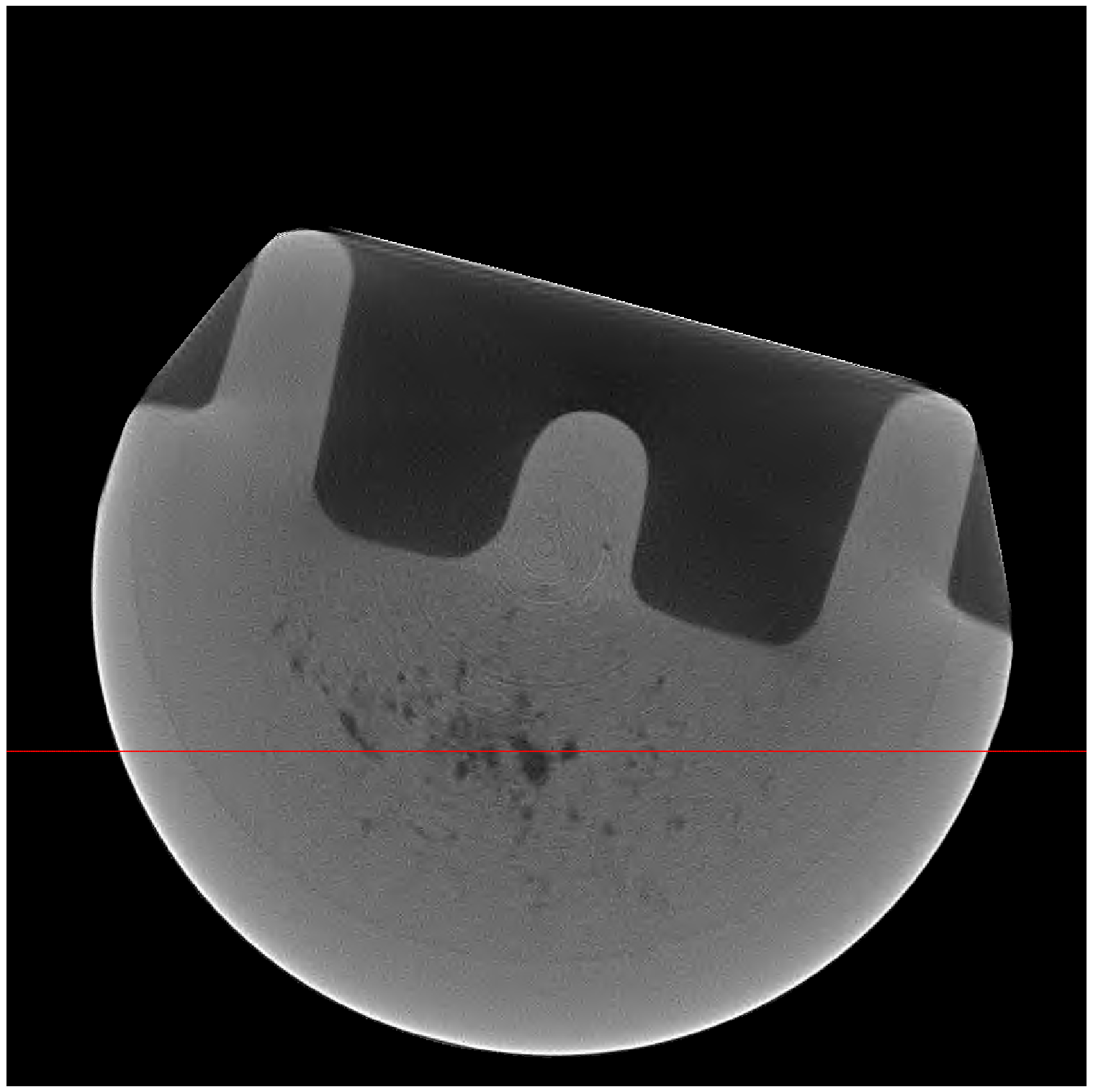}
\label{fig:img_maskDOREfull}
}
\hfil
\subfigure[]{
\includegraphics[width=2.2in]{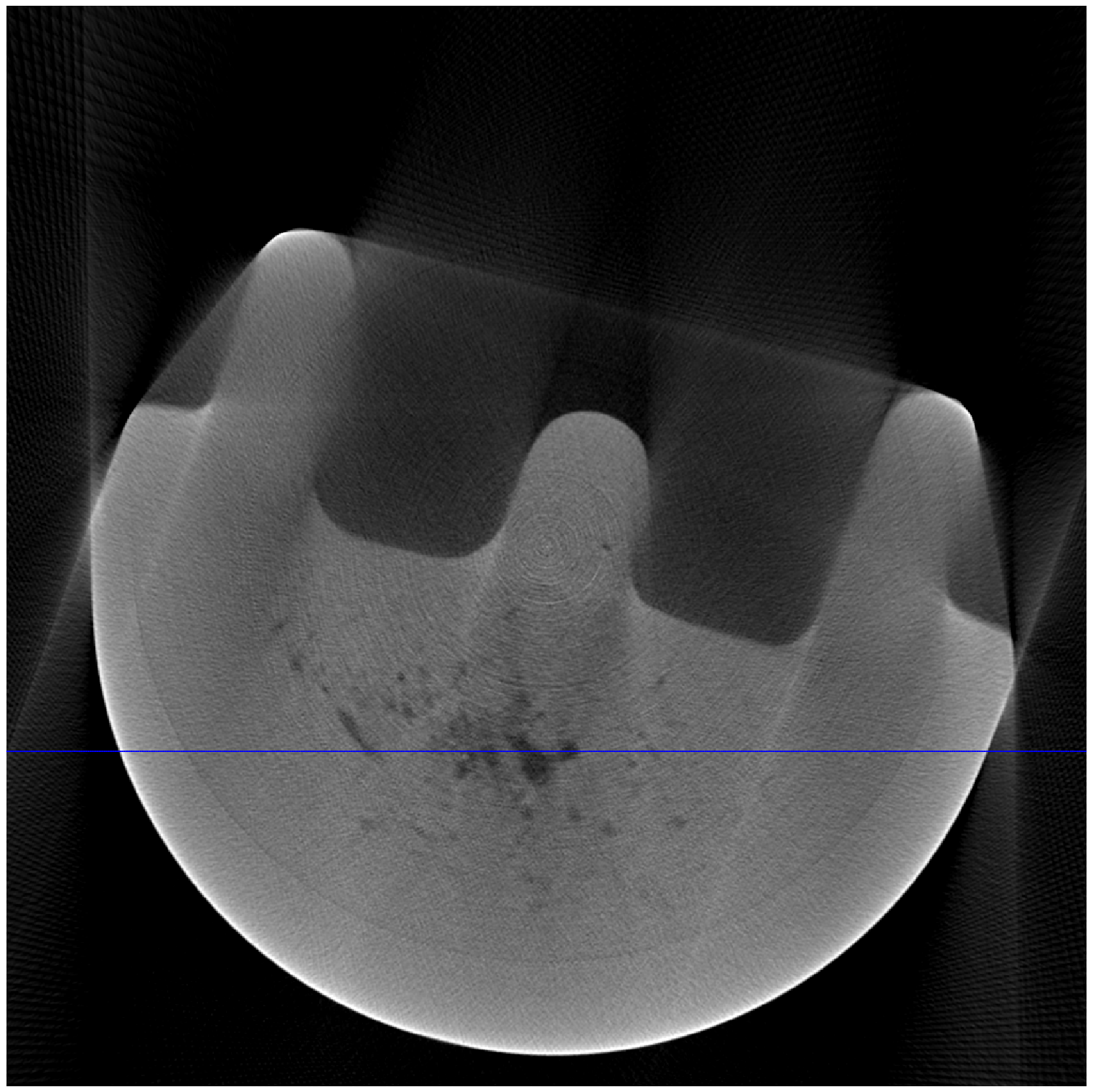}
\label{fig:img_FBPCut20}
}
\hfil
\subfigure[]{
\includegraphics[width=2.2in]{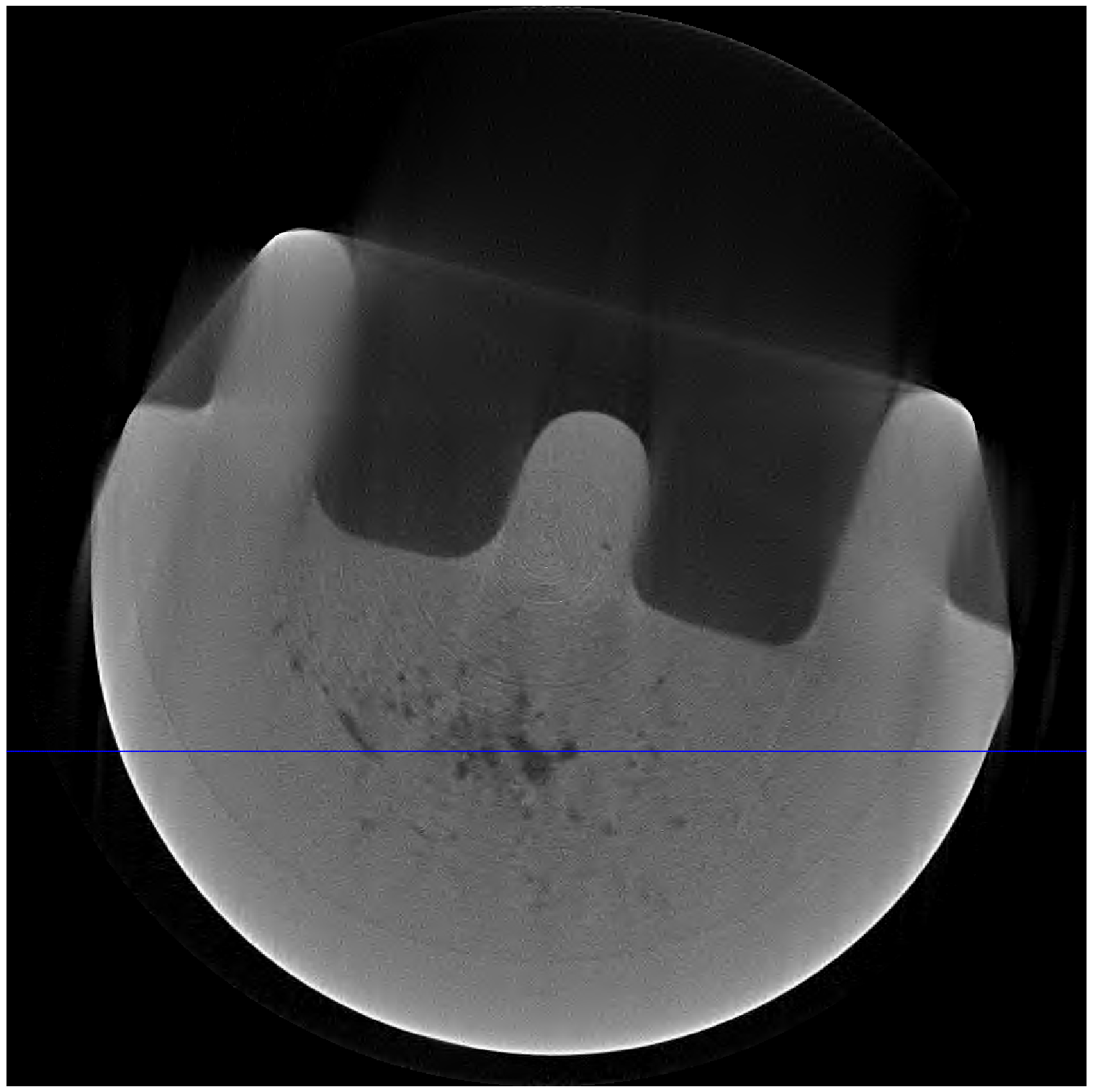}
\label{fig:img_DORECut20}
}
\hfil
\subfigure[]{
\includegraphics[width=2.2in]{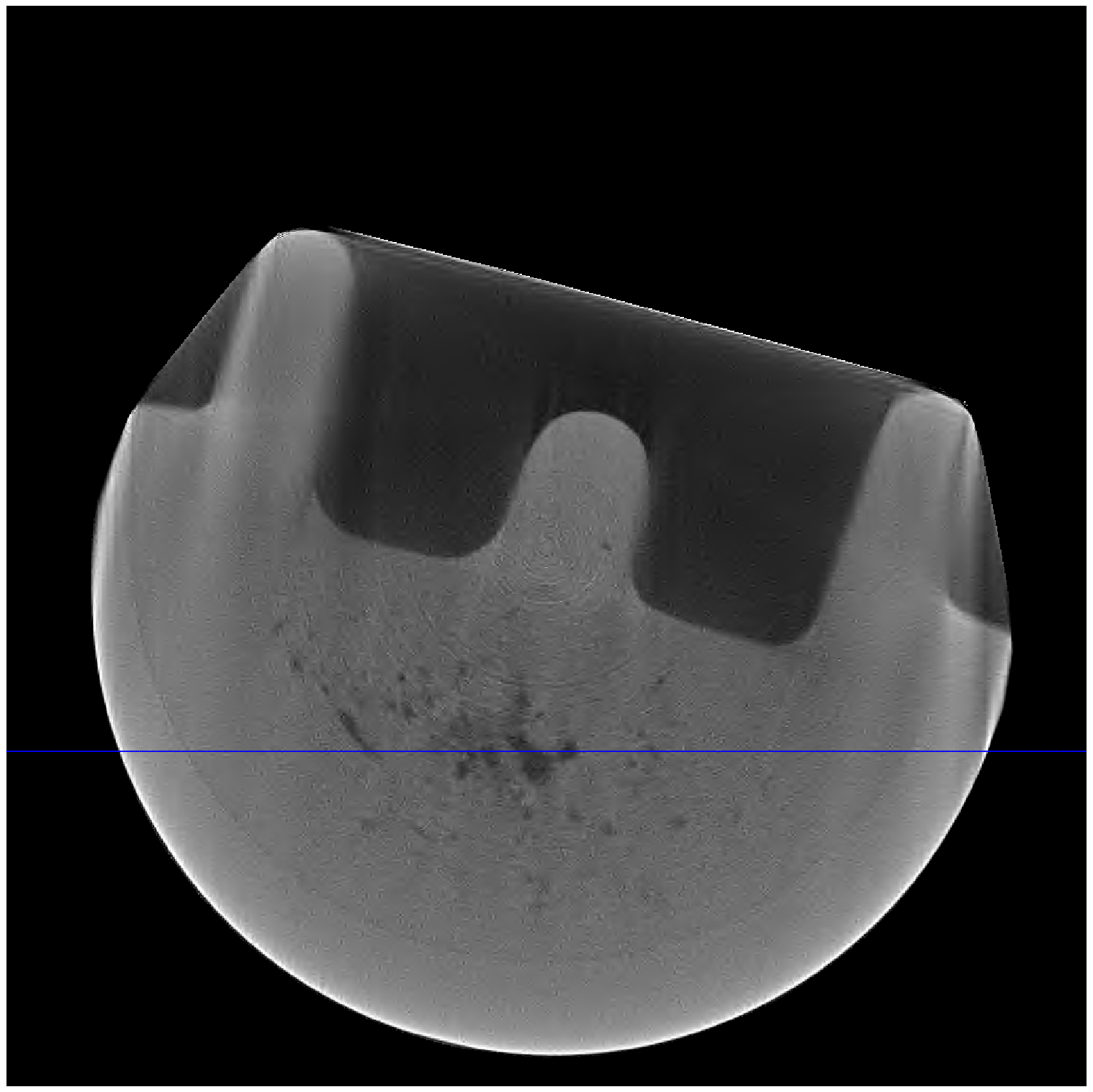}
\label{fig:img_maskDORECut20}
}
\hfil
\subfigure[]{
\includegraphics[width=2.2in]{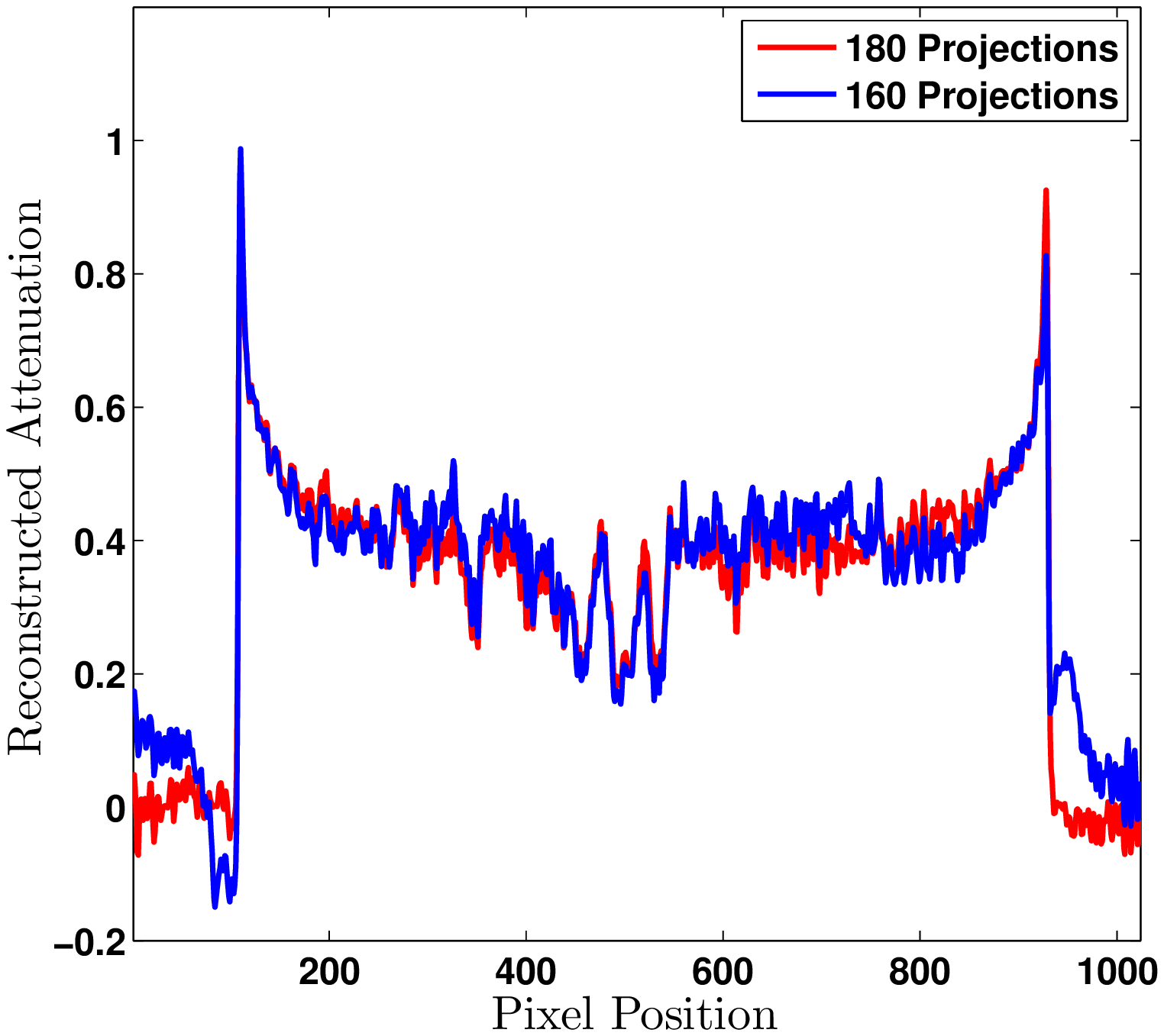}
\label{fig:profile_FBP}
}
\hfil
\subfigure[]{
\includegraphics[width=2.2in]{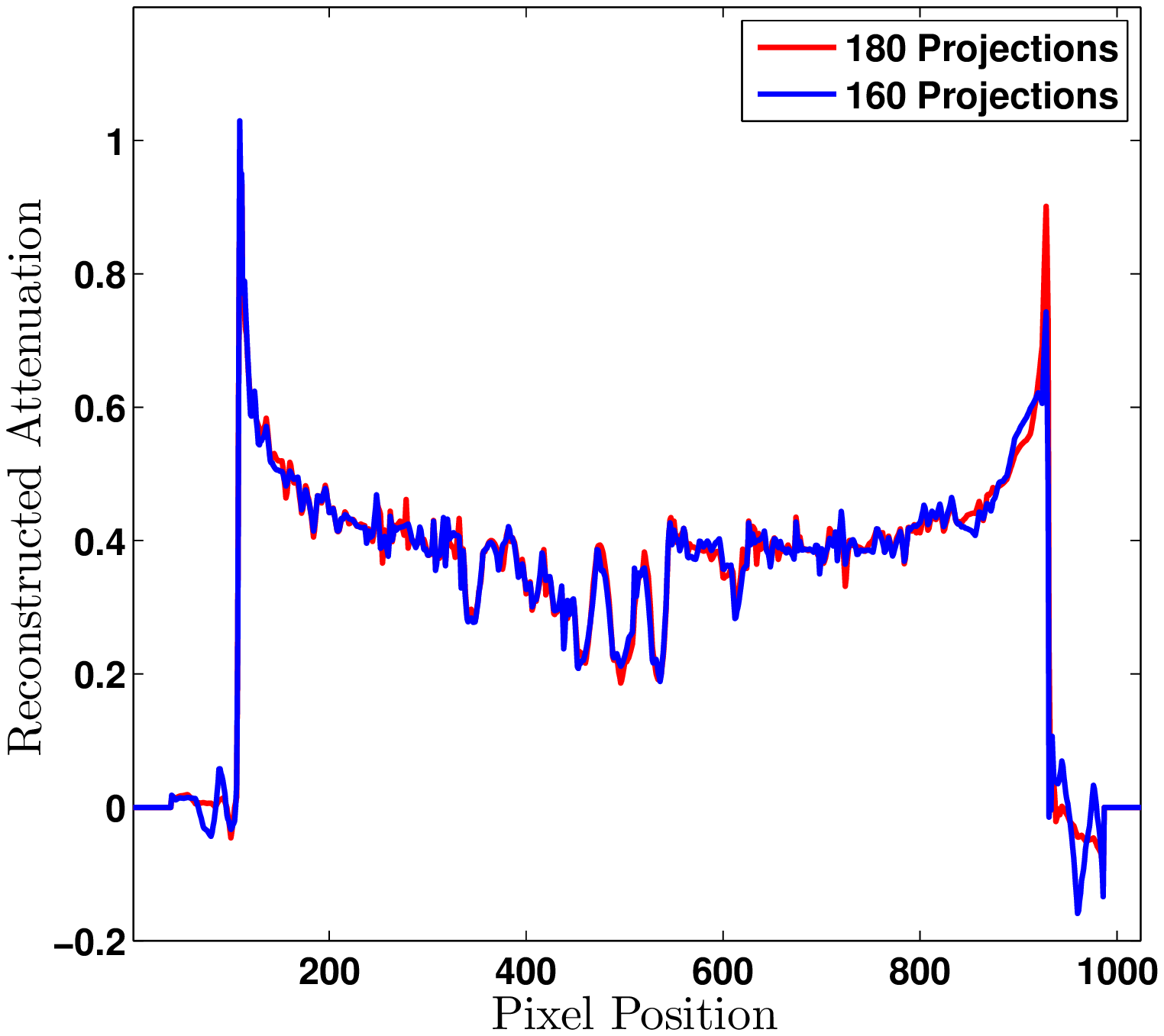}
\label{fig:profile_DORE}
}
\hfil
\subfigure[]{
\includegraphics[width=2.2in]{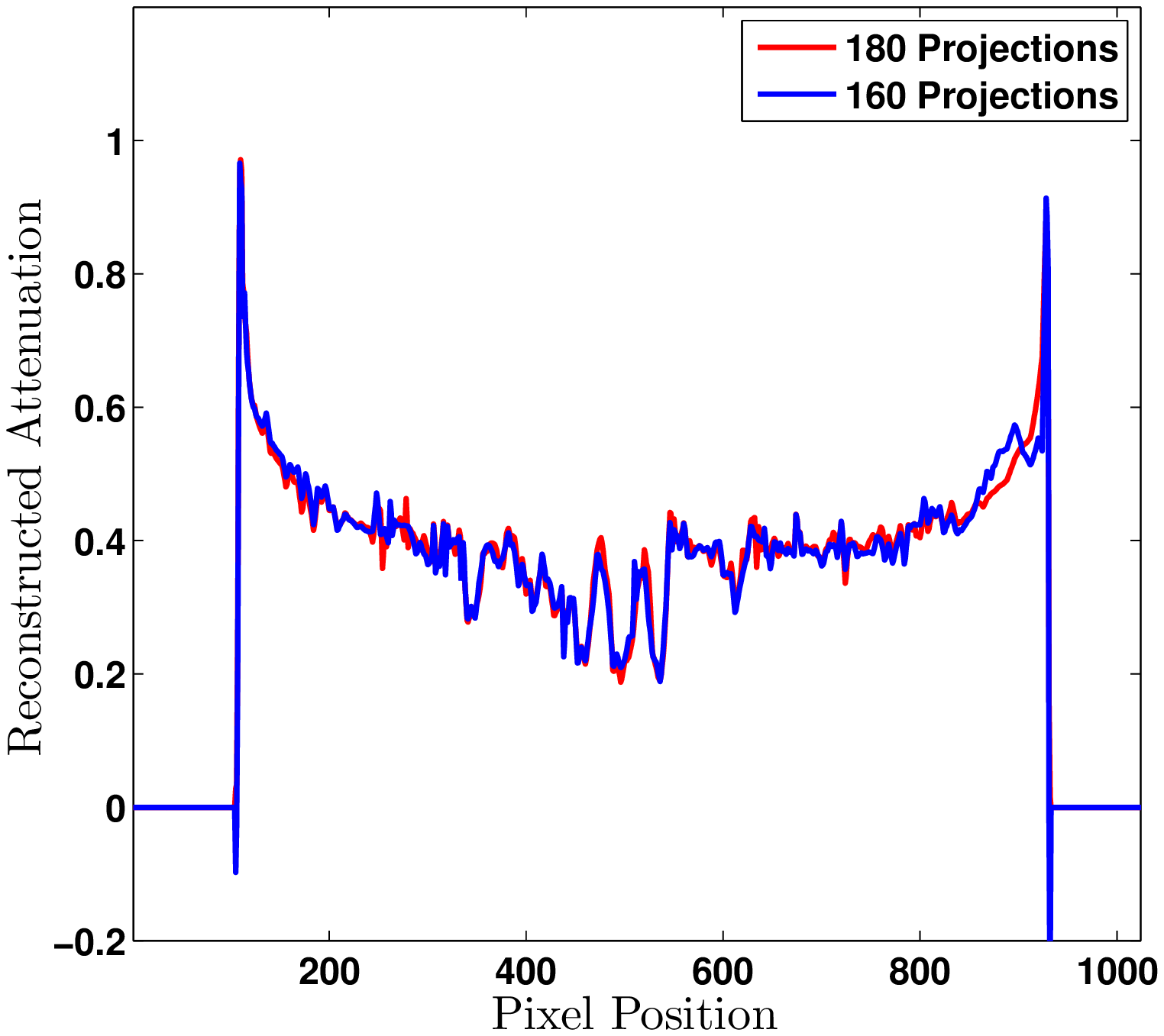}
\label{fig:profile_maskDORE}
}

\caption{FBP, DORE, and mask DORE
reconstructions from (a)--(c) $180$
  projections and (d)--(f) $160$ limited-angle projections; (g)--(i) the
corresponding FBP, DORE, and mask DORE reconstruction 
profiles for slices depicted in (a)--(f).}
\label{fig:numex2}
\end{figure*}

The $m \times m$ orthonormal sparsifying matrix $\Psiit$ is
constructed using the inverse Daubechies-6 DWT matrix.

We consider two measurement scenarios: no missing angles, i.e.\ all $180$
projections available, and limited-angle projections with missing
angle span of $20^\circ$, i.e.\ $160$ projections available.

We compare the reconstruction performances of
mask DORE ($r=15000$) and DORE ($r=20000$) with
  $\epsilon=10^{-8}$;
the mask FPC$_{\footnotesize \mbox{AS}}$ and
  FPC$_{\footnotesize \mbox{AS}}$ schemes
using the
  regularization parameter $\tau=10^{-6} \, \| H^T \, \by
  \|_{\infty}$;
the standard FBP method.
The reconstructions of mask FPC$_{\footnotesize \mbox{AS}}$ and
FPC$_{\footnotesize \mbox{AS}}$ are very similar to those of mask DORE
and DORE; hence we present only the mask DORE and DORE reconstructions
in this example.
Figs.~\ref{fig:img_FBPfull}--\ref{fig:img_maskDOREfull} show the
reconstructions of the FBP, DORE, and mask DORE methods from $180$
projections whereas
Figs.~\ref{fig:img_FBPCut20}--\ref{fig:img_maskDORECut20} show the
corresponding reconstructions from $160$ limited-angle projections.
Figs.~\ref{fig:profile_FBP}--\ref{fig:profile_maskDORE} show the
corresponding reconstruction profiles for slices depicted in
Figs.~\ref{fig:img_FBPfull}--\ref{fig:img_maskDORECut20}.  Observe the
aliasing correction and denoising achieved by the sparse
reconstruction methods.

\appendices

\section*{Appendix}

\renewcommand
    {\theequation}
    {A\arabic{equation}}

\setcounter{equation}{0}

We now prove Theorem~\ref{theorem}. Consider the inequality:
\begin{subequations}
\label{eq:thm1proof3}
\begin{IEEEeqnarray}{rCl}
\IEEEeqnarraymulticol{3}{l}{
\label{eq:thm1proof3_1}
\| \by \kern-0.05em - \kern-0.05em H  \bs_{\sI}^{(q)} \|_2^2 
\kern-0.05em - \kern-0.05em \| \by \kern-0.05em - \kern-0.05em H  \bshat_{\sI}\|_2^2 \kern-0.05em
= \kern-0.05em \| \by \kern-0.05em - \kern-0.05em H  \bs_{\sI}^{(q)} \|_2^2
\kern-0.05em - \kern-0.05em \| \by \kern-0.05em - \kern-0.05em H  \bshat_{\sI}\|_2^2}
\nonumber
\\
& &
    + \: \frac{1}{\mu^{(q)}}\| \bs_{\sI}^{(q)} - \bs_{\sI}^{(q)} \|_2^2
 - \| H \, (\bs_{\sI}^{(q)} - {\bs_{\sI}}^{(q)})\|_2^2
    \nonumber \\ \;\;
&\geq&\| \by - H \, \bshat_{\sI}\|_2^2
    + \frac{1}{\mu^{(q)}}\| \bshat_{\sI} - {\bs_{\sI}}^{(q)} \|_2^2
   -  \| H \, (\bshat_{\sI} - {\bs_{\sI}}^{(q)})\|_2^2
\nonumber
\\ & &
    - \: \| \by - H \, \bshat_{\sI}\|_2^2\\
&=&
\frac{1}{\mu^{(q)}} \, \| \bshat_{\sI} - {\bs_{\sI}}^{(q)} \|_2^2 
- \| H (\bshat_{\sI} - \bs_{\sI}^{(q)}) \|_2^2 \nonumber
\\
\label{eq:thm1proof3_2}
&\ge& (\frac{1}{\mu^{(q)}} - \rho_{\sH}^2) \, \| \bshat_{\sI} - \bs_{\sI}^{(q)} 
\|_2^2
\end{IEEEeqnarray}
\end{subequations}
where (\ref{eq:thm1proof3_1}) follows by using the fact $\bshat_{\sI}$ in 
(\ref{eq:shat}) minimizes
\begin{equation}
\label{eq:thm1proof2}
\mu^{(q)}\| \by - H \, \bs_{\sI} \|_2^2 + \| \bs_{\sI} - {\bs_{\sI}}^{(q)} 
\|_2^2 - \mu^{(q)}\| H (\bs_{\sI} - {\bs_{\sI}}^{(q)})\|_2^2
\end{equation}
over all 
$\bs_{\sI} \in {\cal S}_r$, see also (\ref{eq:signalparameterspace}).
To see this, observe that (\ref{eq:thm1proof2}) can be written as
\begin{equation}
  \| \bs_{\sI} - \bs_{\sI}^{(q)} -  \mu^{(q)} \, H^T \, ( \by - H \, \bs_{\sI}^{(q)} 
  ) \|_2^2 + \mbox{const}
\end{equation}
where const denotes terms that are not functions of $\bs_{\sI}$.
Finally, (\ref{eq:thm1proof3_2}) follows by using
the Rayleigh-quotient property
\cite[Theorem 21.5.6]{Harville}:
$\| H \, (\bshat_{\sI} - {\bs_{\sI}}^{(q)}) \|_2^2 / \| \bshat_{\sI} - 
{\bs_{\sI}}^{(q)} \|_2^2 \leq \rho_{\sH}^2$.
Therefore, in each iteration, $\| \by - H \, \bs_{\sI}^{(q)} \|_2^2$
is guaranteed to not increase if the condition
(\ref{eq:sufficientconditionformu}) holds. Since the sequence $\| \by
- H \, \bs_{\sI}^{(q)} \|_2^2$ is monotonically non-increasing and
lower bounded by zero, it converges to a limit.

\bibliography{IEEEabrv,corresp}   
\bibliographystyle{IEEEtran}   

\end{document}